\documentclass[10pt,twocolumn,letterpaper]{article}

\usepackage{cvpr}              %
\usepackage[accsupp]{axessibility} 
\usepackage{graphicx}
\usepackage{amsmath}
\usepackage{amssymb}
\usepackage{booktabs}

\usepackage{xcolor}
\usepackage{array}
\usepackage{amsmath}

\usepackage{epsfig}
\usepackage{epstopdf}
\usepackage{booktabs}
\usepackage{algpseudocode} 
\usepackage{lipsum}
\usepackage{multirow}
\usepackage{textcomp}
\usepackage{gensymb}
\usepackage{pifont}

\newcommand{\CHD}[1]{{\color{black}#1}}

\usepackage[pagebackref,breaklinks,colorlinks]{hyperref}

\usepackage[capitalize]{cleveref}
\crefname{section}{Sec.}{Secs.}
\Crefname{section}{Section}{Sections}
\Crefname{table}{Table}{Tables}
\crefname{table}{Tab.}{Tabs.}

\def\cvprPaperID{6} %
\def\confName{CVPR}
\def\confYear{2023}

\begin{document}

\def\papertitle{CAT-NeRF: Constancy-Aware Tx$^2$Former for Dynamic Body Modeling}

\title{\papertitle}

\author{Haidong Zhu\ \ \ \ \ \ Zhaoheng Zheng\ \ \ \ \ \ Wanrong Zheng\ \ \ \ \ \ Ram Nevatia\\
University of Southern California\\
{\tt\small \{haidongz|zhaoheng.zheng|wanrongz|nevatia@usc.edu\}}
}
\maketitle

\begin{abstract}
This paper addresses the problem of human rendering in the video with temporal appearance constancy. Reconstructing dynamic body shapes with volumetric neural rendering methods, such as NeRF, requires finding the correspondence of the points in the canonical and observation space, which demands understanding human body shape and motion. Some methods use rigid transformation, such as SE(3), which cannot precisely model each frame's unique motion and muscle movements. Others generate the transformation for each frame with a trainable network, such as neural blend weight field or translation vector field, which does not consider the appearance constancy of general body shape. In this paper, we propose CAT-NeRF for self-awareness of appearance constancy with Tx$^2$Former, a novel way to combine two Transformer layers, to separate appearance constancy and uniqueness. Appearance constancy models the general shape across the video, and uniqueness models the unique patterns for each frame. We further introduce a novel Covariance Loss to limit the correlation between each pair of appearance uniquenesses to ensure the frame-unique pattern is maximally captured in appearance uniqueness. We assess our method on H36M and ZJU-MoCap and show state-of-the-art performance.

\end{abstract}

\section{Introduction}

Rendering an animated person in a video from a novel viewpoint is helpful for several applications, such as game design and simulation, and involves implicit inference of the 3-D human shape and pose.
High-quality human reconstructions require modeling the detailed appearance of each frame, which are expensive in computation and ignore the constant appearance of the same person in the sequence. 
In addition, when encountering occlusions, framewise reconstruction cannot fill in these occluded patterns with knowledge from current viewpoints, which we can find in other frames as the constant appearance of the same person.

\begin{figure}
    \centering
    \includegraphics[width=\linewidth]{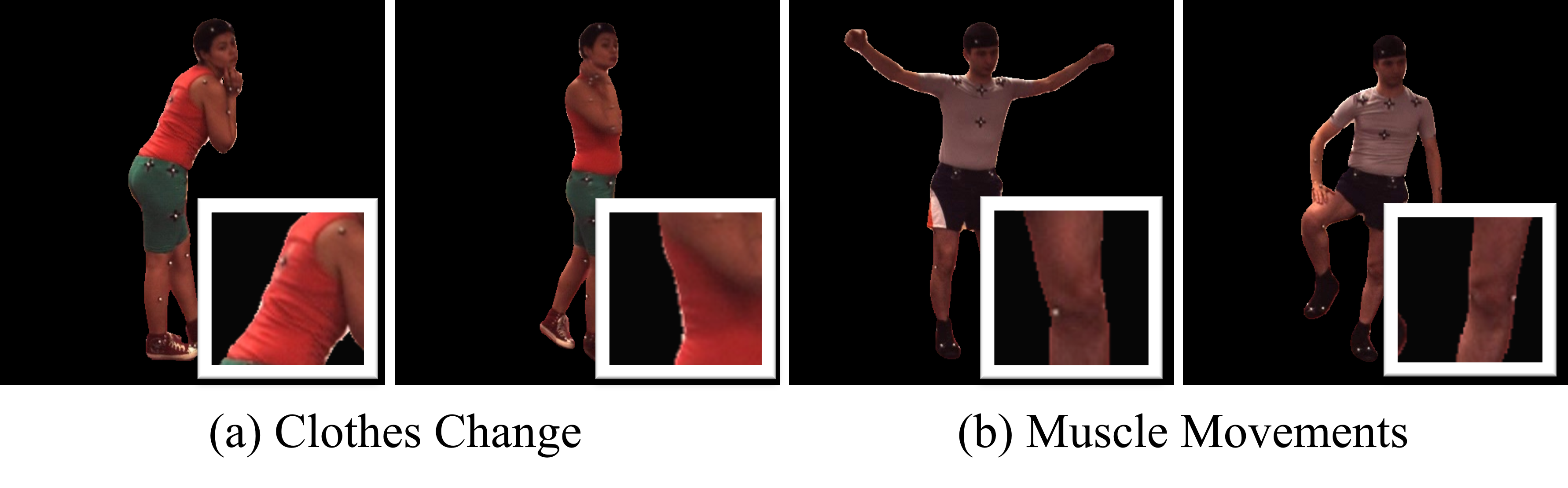}
    \caption{For the dynamic body in the video sequence, we have some unique framewise patterns from (a) change of the patterns on the clothes and (b) change of the muscles in addition to the constant appearances and shapes that are shareable across the frames.}
    \label{fig:example}
\end{figure}

In this work, we focus on mining the appearance constancy among the frames for rendering dynamic body shapes based on the neural radiance field (NeRF) \cite{mildenhall2020nerf}. NeRF implicitly records the density and color of the object, making viewpoint-free rendering possible without requiring explicit geometric modeling. NeRF constructs two spaces for the animation of an object: an observation space for each frame reflecting the observed object shape, and a canonical space for a specific pose shared by all frames. The alignment between these spaces requires understanding both shapes and dynamic motions.

To align the points between these two spaces, researchers use SE(3) \cite{park2021nerfies} or a translation vector field \cite{pumarola2021d,peng2021animatable} for building correspondences between the canonical and observation spaces. The translation vector field predicts the correspondence with a trainable neural network, while SE(3) uses rigid transformation for rotation matrix computation between body parts. 
Considering the non-rigid transformation of the human body shapes, finding such correspondence between these two spaces requires both understanding of rigid transformation for global shapes and non-rigid local movements and motions. We show two examples in Figure~\ref{fig:example}. Based on constant shapes and appearance, dynamic body shapes introduce unique framewise appearance from changing patterns on (a)  the clothes and (b) muscle that cannot be captured with rigid transformation.

To solve this problem, we separate appearance constancy and uniqueness between the frames based on the neural blend weight fields \cite{peng2021animatable} with \textbf{C}onstancy \textbf{A}wareness \textbf{T}x$^2$Former, abbreviated as CAT-NeRF. 
We apply a temporal-constant feature to model the constant appearance shared across all frames and a set of framewise features for each frame to capture the uniqueness. Appearance constancy can help find the missing pattern with the knowledge from other frames when encountering the unseen parts, while appearance uniqueness is to include more frame-specific motions and patterns.

As the temporal-constant and framewise features capture patterns of different levels, the model needs to distinguish the appearance constancy and uniqueness in the feature sequence. We introduce a novel Covariance Loss to minimize the correlation score in the framewise feature. By limiting the similar information shared across the framewise features, these features focus on the frame it is representing and maximally extract the unique patterns in each frame. This can also simultaneously maximize the appearance constancy captured by the temporal-constant feature since the model needs to store these patterns for modeling the dynamic body shapes during optimization. We include more discussion in Sec.~\ref{covariancemat}.

In addition, considering that some appearances are only shared by a small set of frames, directly mining the appearance constancy or uniqueness with Covariance Loss fails to capture these patterns in either level of features.
We introduce Transformer-on-Transformer (Tx$^2$Former), a new way of combining the Transformer layers to fuse the framewise features and focus on useful information based on the current frame. The first Transformer \cite{vaswani2017attention} layer equally combines all  framewise features, and the second Transformer layer takes the feature of the current frame along with the average-pooled output of the first Transformer to select the helpful information for the specific frame. Unlike the typical Transformer that only takes the output of previous layers as input, introducing the feature of the current frame helps the network focus on what needs to be selected from sequential features via self-attention.
We assess our method on two public datasets, ZJU-MoCap \cite{peng2021neural} and H36M \cite{ionescu2013human3}, and show state-of-the-art performance.

In summary, our contributions are as follows: 1) we introduce separating constant and unique appearance for dynamic human body rendering, 2) we introduce CAT-NeRF with a Tx$^2$Former for mining the appearance constancy across the frames and fusing different levels of features across the video, and 3) we introduce a Covariance Loss to mine and preserve unique patterns for each frame.

\section{Related Work}

\textbf{Neural Radiance Field.}
NeRF \cite{mildenhall2020nerf} introduces 2-D images from different viewpoints to reconstruct a cubic neural radiance field for storing the RGB color value and density for each point in the cube. 
Recently some papers \cite{pumarola2021d,park2021nerfies,peng2021animatable,peng2021neural,liu2021neuralactor,noguchi2021neural} introduce decomposing the neural radiance from the observation space to canonical space for modeling the movement of an object. By predicting the correspondences in the canonical space \cite{pumarola2021d,park2021nerfies,peng2021animatable}, deformable NeRF finds the connection between every point in the observation space and the canonical space and uses the moving object in the video for the construction of a unique object. Nerfies \cite{park2021nerfies} and Neural 3-D video synthesis \cite{li2021neural} construct a framewise deformation field for aligning the points in different scenes. HyperNeRF \cite{park2021hypernerf} builds the hyper-space for recording the topological changes. Although these methods perform well on scene reconstruction, it is difficult to apply directly for the animatable body shapes since they heavily rely on the memory of the projection.%

\textbf{Body Shape Reconstruction and Animation.} 
Constructing the human body shape requires complicated hardware by most methods \cite{collet2015high,debevec2000acquiring,dou2016fusion4d,guo2019relightables,su2020robustfusion}. Recently, researchers have mainly developed two different methods: statistic-based methods \cite{bogo2016keep,osman2020star,romero2017embodied,dong2020motion,jiang2020coherent,kanazawa2018end,hedman2021baking,zhuopen,zhu2023gait} and data-based methods \cite{saito2019pifu,saito2020pifuhd,niemeyer2020differentiable,yariv2020multiview,sitzmann2019scene,natsume2019siclope,zheng2019deephuman}. Statistic methods use a predefined body shape with a default linear skinned model for human shape reconstruction. 
Recent work has also introduced some non-statistics models based on the body shape reconstruction for 3-D human body shape reconstruction. \cite{saito2019pifu,saito2020pifuhd} introduce using implicit functions for 3-D estimation. \cite{xiu2021icon} introduced using normals to correct the reconstruction model generated with SMPL shape. With the development of NeRF \cite{mildenhall2020nerf}, researchers also introduced using the implicit function \cite{noguchi2021neural,liu2021neuralactor} and other 3-D representations \cite{shysheya2019textured,aliev2020neural,thies2019deferred,wu2020multi,yoon2021pose} for modeling the static body shapes.

To animate the body shapes and render them in the scene, animatable NeRF, different from the deformable methods, projects the body shape into a canonical space \cite{peng2021animatable,peng2021neural,weng2022humannerf,li2022tava} or a common shape shapes \cite{raj2021anr,prokudin2021smplpix} for projecting the human body shape from different frames into a shared shape or space. Recently, researchers have used statistical methods such as SMPL \cite{loper2015smpl} and SMPL-X \cite{pavlakos2019expressive} representations for body templates. However, these statistical skinned models cannot precisely model the body shapes in the scene, considering different body shapes and clothes. Researchers have proposed different methods to bridge the gap between these two models. Methods such as NARF \cite{noguchi2021neural} and A-NeRF \cite{su2021nerf} do not make explicit canonical space modeling. SNARF \cite{chen2021snarf} and Animatable NeRF \cite{peng2021animatable} utilize the neural blend weight field with frame-level correction for building such correspondence with statistical methods, while TAVA \cite{li2022tava} uses the linear blend skinning (LBS) and apply a constant change for modeling muscles and clothing dynamics. No existing methods take both unique framewise dynamics and temporal constancy into consideration.

\begin{figure*}[t]
    \centering
    \includegraphics[width=0.96\textwidth]{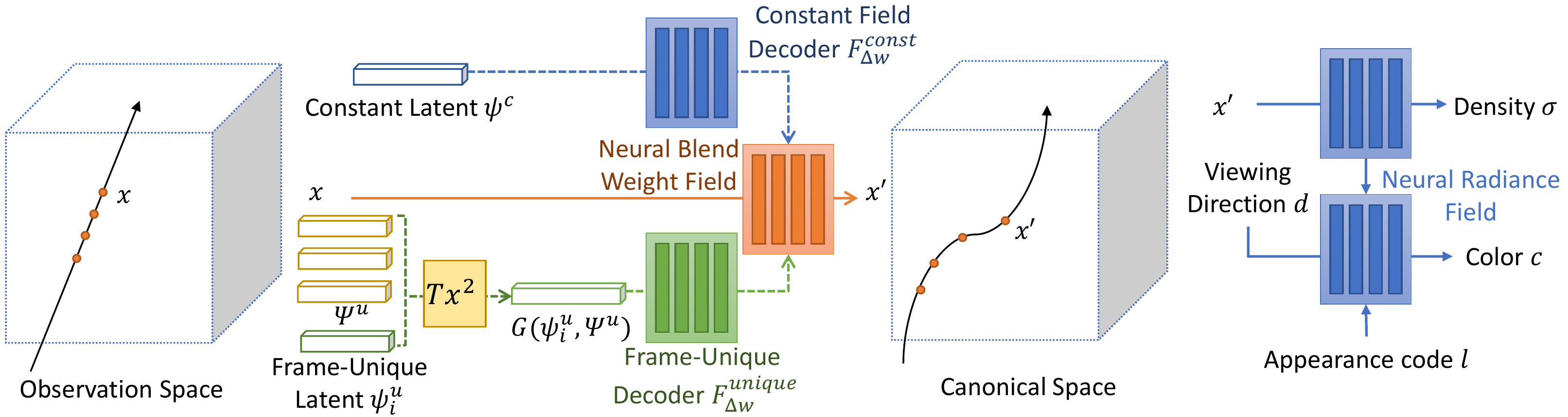}
    \caption{Architecture of the proposed CAT-NeRF for frame uniqueness and temporal constancy decomposition. Rectangles in the boxes are trainable MLP layers. We apply the constant and frame-unique field decoders for temporal constant and unique features, respectively.}
    \label{fig:network}
\end{figure*}

\section{Method}
For each person in a video, we have $n$ camera viewpoints ($n \geq 1$) recording the same sequence from $n$ synchronized and geometry calibrated cameras and generate a frame sequence $\{v_i\}_{i=1,2,...,N}$. $i$ is the current frame number, and $N$ is the length of the sequence. We show the architecture of CAT-NeRF in Figure~\ref{fig:network}. During training, we have both camera positions and corresponding groundtruth images, while during inference, with novel camera viewpoints, we render the image with the corresponding feature vectors by predicting the value of each pixel. 

We first briefly review Animatable NeRF \cite{peng2021animatable} in \ref{aninerf}. We then introduce CAT-NeRF in Sec.~\ref{temporalaware}. For framewise uniqueness, we introduce the covariance loss in Sec.~\ref{covariancemat}, followed by the overall objective for training in Sec.~\ref{obj}.

\subsection{Animatable NeRF for Human Modeling}\label{aninerf}
To represent a dynamic scene or object in the video, Animatable NeRF \cite{peng2021animatable} constructs two spaces: one observation space representing the shape we observe for each individual frame and one canonical space shared by all the frames describing the same object with a default pose. 
For a point $x$ in the observation space, it constructs the scene or object of a frame within a video with two fields representing the density value $\sigma(x)$ and RGB value $c(x)$ of each point following
\begin{equation}
\begin{aligned}
    \sigma(x), z_i(x) &= F_{\sigma}(\gamma_{x}( T_{oc}(x')));\\
    c(x) &= F_c(z_i(x),\ \gamma_{d}(d),\ l))
\end{aligned}
\end{equation}
where $x'$ is the point in the canonical space corresponding to $x$. $d$ is the observation direction and $l$ is the specific feature representative for each frame. $z$ is a learnable representation. $\gamma_{d}$ and $\gamma_{x}$ are the two position encoding functions following \cite{noguchi2021neural}. $T_{oc}$ is the transformation function to find the corresponding point $x$ in the observation space from the point $x'$ in the canonical space, which is formulated as the neural blend weight field \cite{huang2020arch,bhatnagar2020loopreg}. By separating the human body shape with $K$ parts based on the linear skinned model, Animatable NeRF builds the function $T_{oc}$ as 
\begin{equation}
    T_{oc}(x') =(\sum_{k=1}^Kw(x')_kG_k) x'
\end{equation}
where $G_k$ is the $SE (3)$ transformation matrix for the corresponding body part and $w(v)$ is the weight for each point $x'$ in the canonical space. In addition, Animatable NeRF utilizes a frame-wise latent code $\psi_i$ for bridging the differences between the statistic model $w_s(x', S_i)$ and $w_i(x')$ for frame $i$. The final blend weight is calculated as
\begin{equation}
    w_i(x') = norm(F_{\Delta w}(x', \psi_i) + w_s(x', S_i))
\label{eq:baseline}
\end{equation}
where $w_s(\cdot)$ is from the statistical shape $S_i$. %

\subsection{CAT-NeRF: Constant and Unique Appearance}\label{temporalaware}
Although $F_{\Delta w}(x', \psi_i)$ helps bridge the differences between the statistic model and the rendered shape,  frame-wise corrections only use information from current frames and do not utilize information from the whole sequence. Modeling the appearance of dynamic shapes needs to consider both constant appearances that are shared by all the frames and the framewise unique appearances for the unique dynamics that only appear in one or a few frames. 

\begin{figure}
    \centering
    \includegraphics[width=\linewidth]{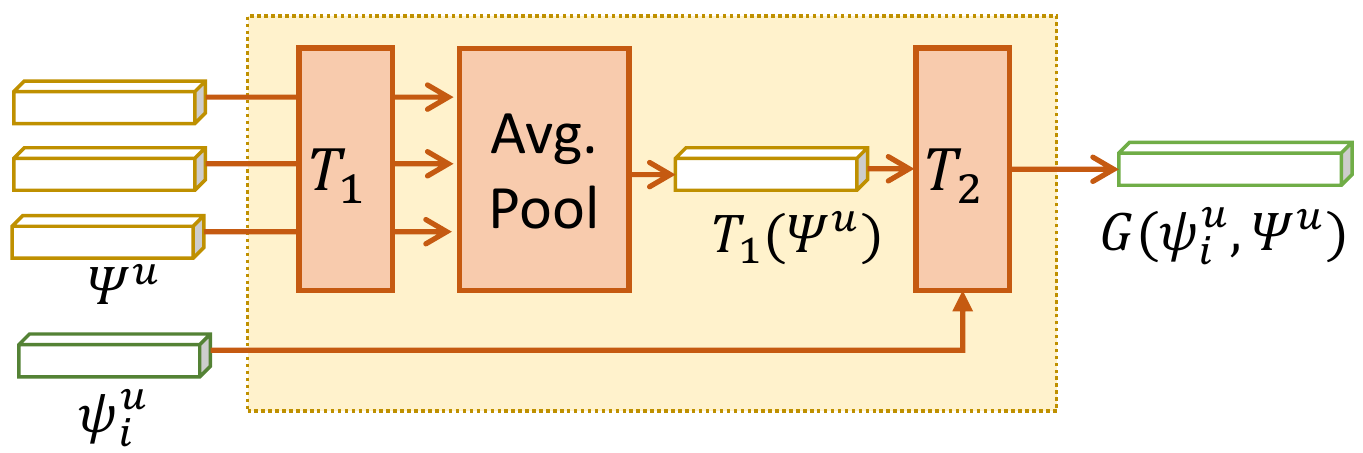}
    \caption{With the collection of frame-unique latent $\Psi^u$ and current frame $\psi_i^u$, we use $T_1$ for fusing the features in the neighbor frames and $T_2$  for combining the features with the current frame.}
    \label{fig:graphical}
\end{figure}

Based on neural blend weight fields, we introduce CAT-NeRF to deal with the constant and unique appearance patterns via mining the temporal constancy in the sequence. Specifically, we use $\psi^u_i$ to store framewise uniqueness for frame $i$, and $\psi^c$ for appearance constancy shared among all the frames. In this way, we decompose the dynamic body rendering of the neural blend weight field following
\begin{equation}
    F_{\Delta w}(x') = F_{\Delta w}^{const}(x_i, \psi^c) + F_{\Delta w}^{unique}(x_i, G(\psi^u_i,\Psi^u)) 
\label{eq:main}
\end{equation}
where $F_{\Delta w}^{const}$ and $F_{\Delta w}^{unique}$ represent two networks to decode the corresponding temporal constant and framewise adjustment of the appearance for the dynamic person, respectively. $G$ is the Tx$^2$Former for combining framewise features for frame $i$ as Figure~\ref{fig:graphical}.

\CHD{
Since some appearances of the body are shared among a small set of frames that are neither constant nor framewise unique, we introduce Transformer-on-Transformer (Tx$^2$Former) as a novel way of combining two Transformer \cite{vaswani2017attention} layers.
Tx$^2$Former combines the framewise-unique features to model the appearances only shared by a set of frames. 
We have two stacks of Transformer layers \cite{vaswani2017attention} to fuse the framewise feature for frame $i$ with other frames in the video. The final output of $G(\psi^u_i,\Psi^u)$ follows
\begin{equation}
    G(\psi^u_i,\Psi^u) = T_2(\psi^u_i, T_1(\Psi^u))
\end{equation}
where $T_1$ and $T_2$ are two stacks of Transformer encoder layers.
The first Transformer encoder fuses all the unique features to generate a global understanding of what is included for all the frames in this sequence besides appearance constancy $\psi^c$. We average pool the output $T_1$ and concatenate it with the corresponding framewise-unique shape $\psi^u_i$ of the current frame before sending it to $T_2$.

\textbf{Discussion:} 
Unlike other vision transformers \cite{dosovitskiy2020image,touvron2021training,liu2021swin} that require the output of each patch used in the input, the reconstruction of the current frame relies more on $\psi^u_i$ and compared with the output of the other frames. In addition, using the original Transformer generates $N-1$ outputs that are not used for rendering since we only focus on the output features for frame~$i$. Moreover, traditional Transformers compute the self-attention across each pair of frames, which still focuses on the frame-level understanding without sequential knowledge. In Tx$^2$Former, concatenating the pooled features and $\psi^u_i$ shortens the sequence length for the input of $T_2$, making the model focus on $\psi^u_i$ and select the sequence-level information from $T_1$'s output based on $\psi^u_i$.
}

\subsection{Covariance Loss for Unique Patterns}\label{covariancemat}
To split the latent representation for framewise uniqueness $\psi^u_i$ and the constancy $\psi^c$ shared by all frames, we introduce a covariance-related loss to separate the information stored in these two representations in individual frames. For a video $i\in \{1,2,...,N\}$, where $N$ is the overall number of frames, we have the collection of frame unique representations $\Psi^u = (\psi_1^u,\psi_2^u,...,\psi_n^u)$ for framewise uniqueness. The covariance loss is as follows
\begin{equation}
    L_{cov} = \frac{\sum ||cov(\Psi^u, \Psi^u)|| - \sum Diag(||cov(\Psi^u, \Psi^u)||)}{(N-1)^2}
    \label{eq:covloss}
\end{equation}
$cov(\cdot)$ is the covariance matrix for $\Psi^u$, and $\sum$ represents the sum of every element in its bracket. $||\cdot||$ indicates using the absolute value for every element. 

For a covariance matrix, each element $cov_{i,j}$ represents the covariance value between the two features $\psi_i$ and $\psi_j$. Since the diagonal value for the covariance matrix is the covariance between a variable to itself, it is the variance of this variable. Since 0 variance indicates all the elements in the vector are 0, we remove this item in our loss function. 

\textbf{Discussion:} 
When the covariance value between two variables is not zero, these two variables tend to vary in the same or opposite direction. Since the range for the covariance value is in $(-\infty, +\infty)$, we use the absolute value for each element to change its range to $[0,+\infty)$. A smaller absolute value for covariance indicates that these two variables are less likely to vary together. In this case, we set 0 as the minimum of the loss function and our target for optimization. 
By reducing the correlation between each pair of the framewise-unique features, $\psi_i^l$ for frame $i$ is less likely to vary along with other frames and increasingly represent the uniqueness of each frame. This also allows the constant feature $\psi^c$ to capture the maximum amount of the temporal constancy information and reduce the reuse of the same feature representations in $\psi_i^u$. 

To show that $L_{cov}$ can minimize temporal constancy information in $\psi^u$ across frames, for each feature $\psi_i$, we can decompose it into two vectors: $a$, which is related to other feature $\psi_j$ at timestamp $j$, $i\neq j$, and $b$, which is unique and not shared with other features. In CAT-NeRF, we use the learnable feature vector $\psi^u$ to represent the frame's unique feature and $\psi^c$ to represent the constancy. Thus we have 
$$\psi_i = \psi^u + \psi^c = a + b$$
If two features are related to each other, the decoded results are similar after decoding with the same network and have strong connections. On the contrary, if two features represent two distinct shapes sharing no similarity, the correlation between them should be minimized. Thus by applying the correlation loss on $\psi^u$, we allow the sharable feature $a$ to be minimally captured by $\psi^u$ and make $\psi^c$ learn the constancy. In the meantime, since the constant feature is captured by $\psi^c$, $\psi^u$ has more capability of capturing the framewise-unique feature $b$ and includes more fine-grained details for each frame in the sequence.

\subsection{Objective}\label{obj}
To train the model, we follow \cite{peng2021animatable} to build the objective function for training $\psi^c$, $\psi^u_i$, $F_{\Delta w}^{const}$, $F_{\Delta w}^{unique}$, $G(\cdot)$, $F_{\sigma}$ and $F_c$ jointly. The final objective for training is
\begin{equation}
\begin{aligned}
    L &= L_{cov} + L_{rgb} + L_{nsf}\\
    L_{rgb} &= \sum_{r\in R}|| \Tilde{C}_i{(r)}-C_i(r)||_2\\
    L_{nsf} &= \sum_{x\in \mathcal{X}}|| w_i(x)-w^{can}(T_{oc}(x))||_1
\end{aligned}
\label{eq:finalloss}
\end{equation}
where $R$ is the collection for all the rays that go through the pixel and $\mathcal{X}$ represents all the points sampled in the volumetric field. $||\cdot||_k$ is the $k$-norm value. $L_{rgb}$ assess the differences between the final rendered color $\Tilde{C}_i(r)$ with the groundtruth value $C_i(r)$ for each pixel. Since the blend weight fields between the points in observation space $x$ and canonical space $T_{oc}(x)$ should be the same for Eq.~\ref{eq:baseline}, we follow \cite{peng2021animatable} for establishing $L_{nsf}$ to minimize the difference.

\section{Experiments}
In this section, we present our settings and results. We first show the dataset description in Sec.~\ref{ds}, followed by the implementation details in Sec.~\ref{id}. With these experimental details, we show our results in Sec.~\ref{numr} and \ref{visr}. 

\subsection{Datasets}\label{ds}
In our experiments, we compare our methods with baseline methods on two different datasets: H36M \cite{ionescu2013human3} and ZJU-MoCap \cite{peng2021neural}. These two datasets capture the moving pattern of different poses of the same person from different camera viewpoints whose viewpoints are available. We follow \cite{peng2021animatable} to select the frames and generate the splits for training and inference in our experiment.

\textit{\textbf{H36M}} \cite{ionescu2013human3} includes videos of different poses for the same and unique person from 4 different camera viewpoints. In our experiment, we follow \cite{peng2021animatable} to select the videos from subjects S1, S5, S6, S7, S8, S9 and S11. We use the first three viewpoints (0, 1 and 2) for training and the remaining for inference for the four camera viewpoints. For novel view synthesis, the number of frames for training and testing for these subjects varies between 30 and 60 for each camera viewpoint. For novel pose synthesis, we use 49 to 200 frames for each subject for evaluation. The size of the image is set to $1002 \times 1000$.

\textit{\textbf{ZJU-MoCap}} \cite{peng2021neural} includes videos captured from 21 different cameras to collect different human poses. We follow \cite{peng2021animatable} to use the videos in four categories, “Twirl”, “Taichi”, “Warmup”, and “Punch1” in our experiment. We select four viewpoints from positions 0, 6, 12 and 18 from the dataset for training and use the remaining 17 viewpoints for inference. For novel view synthesis, the number of frames for training and testing for these subjects varies between 60 and 400 for each camera viewpoint. For novel pose synthesis, we use 346 to 1,000 frames for each subject for evaluation. The size of the image for each frame is $1024 \times 1024$.

\begin{table*}[t]
\centering
\def\lw{1}
\def\lf{1}
\def\ls{0.05}
\resizebox{0.9\linewidth}{!}
{
\begin{tabular}{p{1cm}<{\centering}p{\ls cm}p{\lw cm}<{\centering}p{\lw cm}<{\centering}p{\lw cm}<{\centering}p{\lw cm}<{\centering}p{\lw cm}<{\centering}p{\lf cm}<{\centering}p{\ls cm}p{\lw cm}<{\centering}p{\lw cm}<{\centering}p{\lw cm}<{\centering}p{\lw cm}<{\centering}p{\lw cm}<{\centering}p{\lf cm}<{\centering}p{\ls cm}} 
\toprule
\multirow{2}{*}{Splits}  && \multicolumn{6}{c}{PSNR ($\uparrow$)} && \multicolumn{6}{c}{SSIM ($\uparrow$)} \\

\cline{3-8} \cline{10-15}  \\ [-8pt]
&& Rand & NT & NHR & SMPLpix & AN & Ours&& Rand & NT & NHR & SMPLpix & AN & Ours\\
\midrule
S1 && 17.79 & 20.98 & 21.08 & 22.01 & 22.05 & \textbf{24.52} && 0.784 & 0.860 & 0.872 & 0.882 & 0.888 & \textbf{0.905}  \\
S5 && 18.19 & 19.87 & 20.64 & 23.35 & 23.27 & \textbf{24.23} && 0.781 & 0.855 & 0.872 & 0.879 & 0.892 & \textbf{0.901} \\
S6 && 18.08 & 20.18 & 20.40 & 21.09 & 21.13 & \textbf{24.24} && 0.769 & 0.816 & 0.830 & 0.860 & 0.854 & \textbf{0.875} \\
S7 && 16.51 & 20.47 & 20.29 & 22.03 & 22.50 & \textbf{24.11} && 0.753 & 0.856 & 0.868 & 0.888 & 0.890 & \textbf{0.899} \\
S8 && 16.94 & 16.77 & 19.13 & 22.22 & 22.75 & \textbf{23.66} && 0.762 & 0.837 & 0.871 & 0.895 & 0.898 & \textbf{0.904} \\
S9 && 18.26 & 22.96 & 23.04 & 23.99 & 24.72 & \textbf{25.95} && 0.770 & 0.873 & 0.879 & 0.902 & 0.908 & \textbf{0.909} \\
S11&& 18.98 & 21.71 & 21.91 & 22.05 & 24.55 & \textbf{25.26} && 0.756 & 0.859 & 0.871 & 0.889 & 0.902 & \textbf{0.905} \\
\midrule
Average && 17.82 & 20.42 & 20.93 & 22.39 & 23.00 & \textbf{24.57} && 0.768 & 0.851 & 0.866 & 0.885 & 0.890 & \textbf{0.900} \\
\bottomrule
\smallskip
\end{tabular}
}
\caption{Results of novel view synthesis on H36M dataset. NT and AN represents Neural Textures and Animatable NeRF respectively. ($\uparrow$) indicates higher results are better.} 
\label{tab:h36mview}
\end{table*}

\begin{table*}[t]
\centering
\def\lw{1}
\def\lf{1}
\def\ls{0.05}
\resizebox{0.9\linewidth}{!}
{
\begin{tabular}{p{1cm}<{\centering}p{\ls cm}p{\lw cm}<{\centering}p{\lw cm}<{\centering}p{\lw cm}<{\centering}p{\lw cm}<{\centering}p{\lw cm}<{\centering}p{\lf cm}<{\centering}p{\ls cm}p{\lw cm}<{\centering}p{\lw cm}<{\centering}p{\lw cm}<{\centering}p{\lw cm}<{\centering}p{\lw cm}<{\centering}p{\lf cm}<{\centering}p{\ls cm}} 
\toprule
\multirow{2}{*}{Splits}  && \multicolumn{6}{c}{PSNR ($\uparrow$)} && \multicolumn{6}{c}{SSIM ($\uparrow$)} \\

\cline{3-8} \cline{10-15}  \\ [-8pt]
&& Rand & NT & NHR & SMPLpix & AN & Ours&& Rand & NT & NHR & SMPLpix & AN & Ours\\
\midrule
S1 && 16.51 & 20.09 & 20.48 & 21.90 & 21.37 & \textbf{23.34} && 0.741 & 0.837 & 0.853 & 0.875 & 0.868 & \textbf{0.888} \\
S5 && 17.80 & 20.03 & 20.72 & 23.01 & 22.29 & \textbf{23.32} && 0.749 & 0.843 & 0.860 & 0.878 & 0.875 & \textbf{0.888} \\
S6 && 17.94 & 20.42 & 20.47 & 21.89 & 22.69 & \textbf{24.55} && 0.805 & 0.844 & 0.856 & 0.865 & 0.884 & \textbf{0.891} \\
S7 && 15.92 & 20.03 & 19.66 & 22.12 & 22.22 & \textbf{22.72} && 0.723 & 0.838 & 0.852 & 0.873 & \textbf{0.878} & \textbf{0.878}\\
S8 && 16.36 & 16.69 & 18.83 & 22.01 & 21.78 & \textbf{22.90} && 0.750 & 0.824 & 0.855 & 0.889 & 0.882 & \textbf{0.895}\\
S9 && 17.53 & 22.20 & 22.18 & 23.91 & 23.72 & \textbf{24.74} && 0.738 & 0.851 & 0.860 & 0.890 & 0.886 & \textbf{0.892}\\
S11&& 19.64 & 21.72 & 22.12 & 22.45 & 23.91 & \textbf{24.24} && 0.747 & 0.854 & 0.867 & 0.875 & 0.889 & \textbf{0.891}\\
\midrule
Average && 17.39 & 20.17 & 20.64 & 22.47 & 22.55 & \textbf{23.68} && 0.750 & 0.841 & 0.858 & 0.878 & 0.880 & \textbf{0.889}\\
\bottomrule
\smallskip
\end{tabular}
}
\caption{Results of novel pose synthesis on H36M dataset. NT and AN represents Neural Textures and Animatable NeRF respectively.} 
\label{tab:h36mpose}
\end{table*}

\subsection{Implementation Details}\label{id}

\textit{\textbf{Training and Inference.}} To extract the SMPL shapes for RGB frames, we follow \cite{joo2018total} to generate the SMPL reconstruction. We follow \cite{huang2020arch,bhatnagar2020loopreg} to generate the neural blend weight field to find the three nearest points on the skinned model for building the field. For each batch, we sample 4,096 rays and for each ray, we sample 64 points.

To train the network for novel camera viewpoints, we follow \cite{peng2021animatable} to implement our network. For  $\psi^c$ and $\psi^u$, we set the number for both dimensionalities as 128. To train the model for novel poses, we first get a model for a novel view with constant and unique features, respectively. After that, we copy the frame-unique features reconstructed for the current poses to the novel poses and use the smooth-L1 loss for the novel constant feature for body shape and the original feature generated from the novel view. For both training steps, we use the Adam optimizer \cite{kingma2014adam} and set the initial learning rate as $5e-4$ and decay it to $\frac{1}{10}$ after 1,000 epochs with an exponential training scheduler following \cite{peng2021animatable}. The number of epochs is set to 400.
During inference, we use the pretrained constant and unique features to decode the density and RGB value for each point in the field. Our network takes 6-8 hours to train on a novel viewpoint setting and 20-30 hours to train on the novel pose setting for each subject on an Nvidia A40 or A100 GPU.

\textit{\textbf{Metrics.}} For our experiment, we have two different metrics between the projection of new images with our generated results for comparison, PSNR and SSIM \cite{wang2004image}. PSNR and SSIM describe the quality of the reconstructed image to the original image. Higher values represent better performance for both metrics.

\textit{\textbf{Baseline Methods.}} In our experiment, we compare our method with Neural Texture \cite{thies2019deferred}, NHR \cite{wu2020multi} and Animated NeRF \cite{peng2021animatable}. We also report  SMPLpix \cite{prokudin2021smplpix} on H36M, along with NeuralBody \cite{peng2021neural} and HumanNeRF \cite{weng2022humannerf} on ZJU-MoCap, since the authors did not provide numbers on the other datasets. We follow \cite{peng2021animatable} to use the body shape reconstruction from SMPL \cite{loper2015smpl} as the input for Neural Texture \cite{thies2019deferred} as the coarse mesh to render the image and use the points sampled from the SMPL body shape reconstruction as the input for NHR. In addition, we also compare a randomly generated image with untrained Animatable NeRF for PSNR and SSIM since these two scores for randomly generated images are not the lowest theoretical values (0) for both of these two metrics. {For all these methods, we follow the setting of Animatable NeRF\footnote{\label{note1}\url{https://github.com/zju3dv/animatable_nerf}}.}

\subsection{Numerical Results}\label{numr}
We show our numerical results on H36M \cite{ionescu2013human3} and ZJU-MoCap \cite{peng2021neural}, along with the ablation study for $L_{cov}$ and Tx$^2$Former architecture, in this subsection. We present other ablation studies, such as hyperparameter selections, in the \textbf{supplementary materials}.

\textit{\textbf{Results on H36M.}} We show the numerical results for H36M dataset rendered for novel view and novel pose in Table~\ref{tab:h36mview} and Table~\ref{tab:h36mpose} respectively. For the results in the tables, NT is the result for Neural Texture \cite{thies2019deferred} and AN is the result from Animatable NeRF \cite{peng2021animatable}.

For the results on the novel view setting shown in Table~\ref{tab:h36mview}, we outperform the state-of-the-art baseline method, Animatable NeRF \cite{peng2021animatable}, on all splits in the dataset for both metrics we assessed. Since the PSNR and SSIM for a randomly generated image are not 0, we achieve a $30.3\%$ and $8.2\%$ relative improvement compared with the differences between our baseline method, Animatable NeRF, and the randomly generated images. Constant features find the appearance constancy across different frames in the video and make full use of all the features, as well as reduce randomness with the assistance of constant appearances in the other frames, while the framewise feature can refine the details based on the constant appearances.

In addition to the results for novel viewpoints, we show the results on novel poses in Table~\ref{tab:h36mpose} comparing with the baseline methods on both metrics. Our method consistently improves on most of the splits for both metrics. Specifically, compared with Animatable NeRF, the best baseline method in our experiment on H36M, we have $21.9\%$ and $6.9\%$ relative improvements on PSNR and SSIM, respectively.  %

\begin{figure*}[t]
\centering
\def\lw{2.5}
\def\ls{0.05}
\resizebox{0.9\linewidth}{!}
{
\centering
\begin{tabular}{p{\lw cm}<{\centering}p{\lw cm}<{\centering}p{\lw cm}<{\centering}p{\lw cm}<{\centering}p{\lw cm}<{\centering}p{\lw cm}<{\centering}}
        \\
        
        \includegraphics[width=\lw cm]{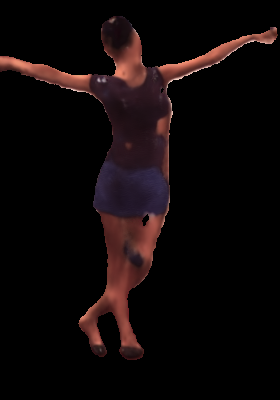}&
        \includegraphics[width=\lw cm]{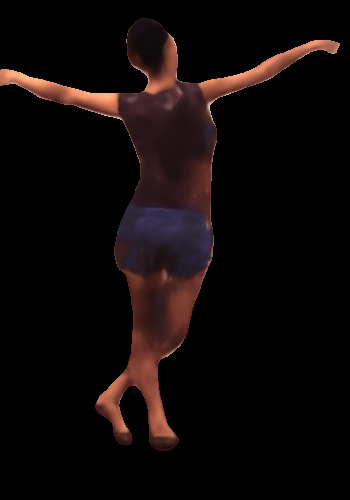}& 
        \includegraphics[width=\lw cm]{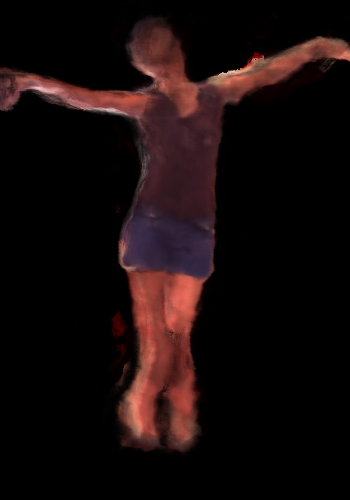}& 
        \includegraphics[width=\lw cm]{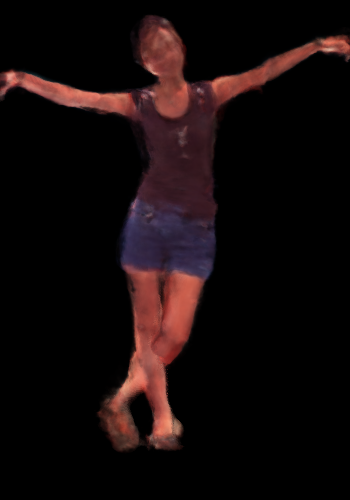}& 
        \includegraphics[width=\lw cm]{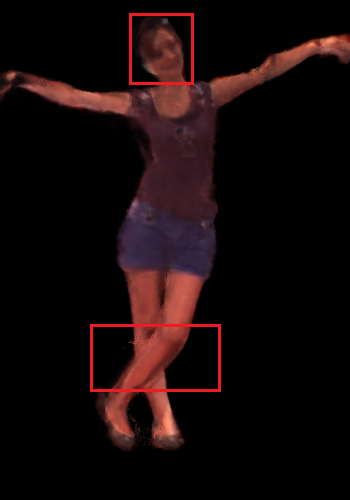}&
        \includegraphics[width=\lw cm]{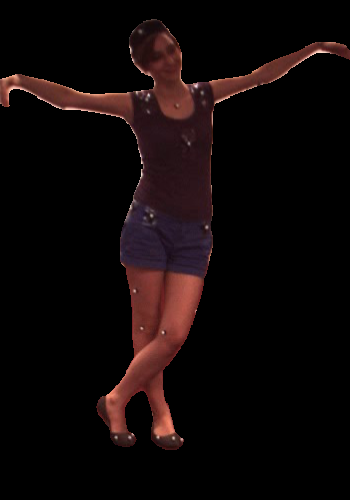}\\
        
        \includegraphics[width=\lw cm]{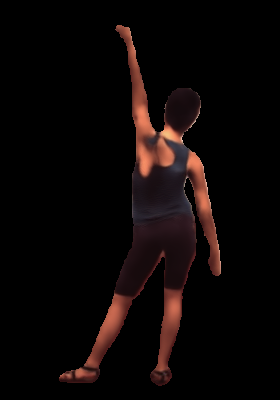}&
        \includegraphics[width=\lw cm]{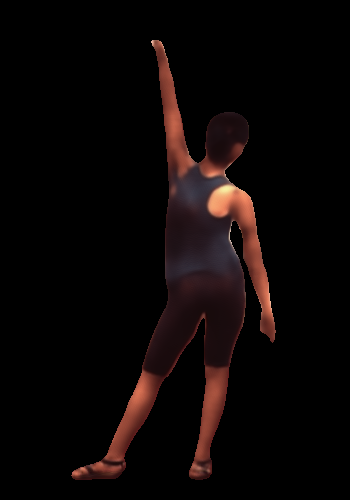}& 
        \includegraphics[width=\lw cm]{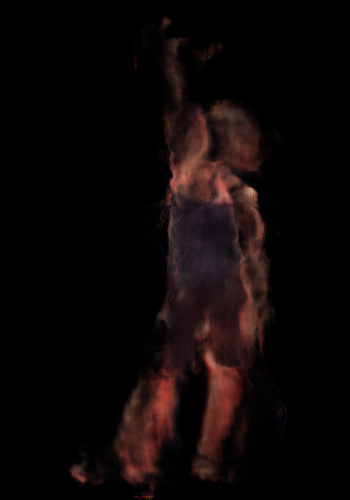}& 
        \includegraphics[width=\lw cm]{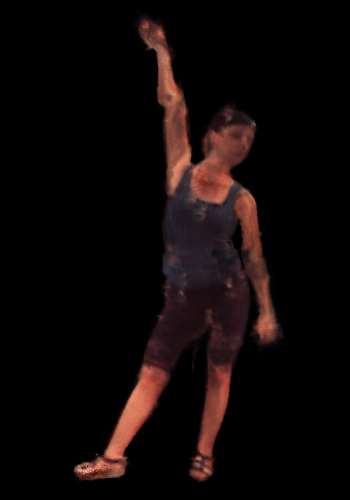}& 
        \includegraphics[width=\lw cm]{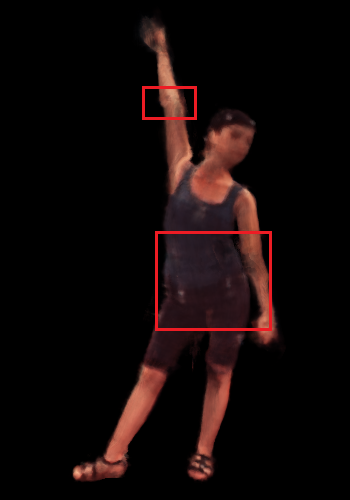}&
        \includegraphics[width=\lw cm]{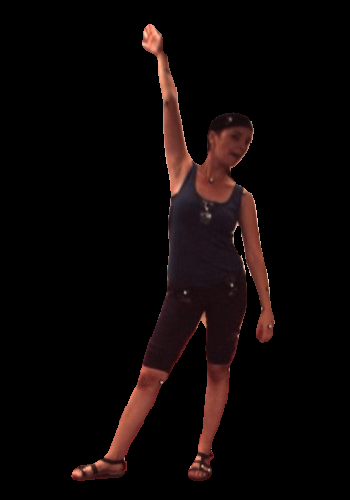}\\
        
        \includegraphics[width=\lw cm]{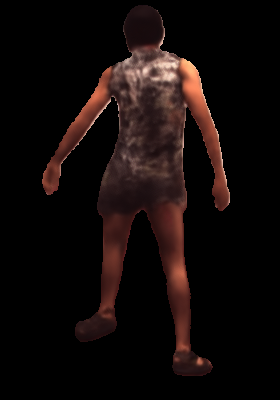}&
        \includegraphics[width=\lw cm]{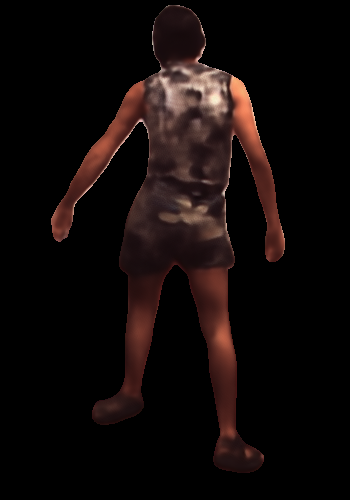}& 
        \includegraphics[width=\lw cm]{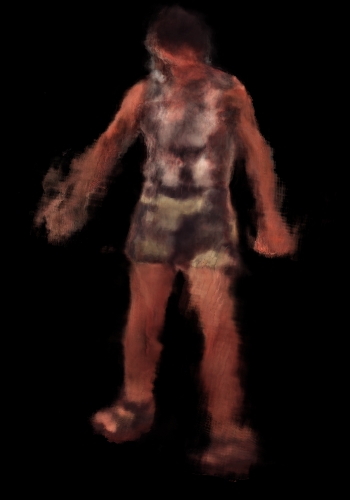}& 
        \includegraphics[width=\lw cm]{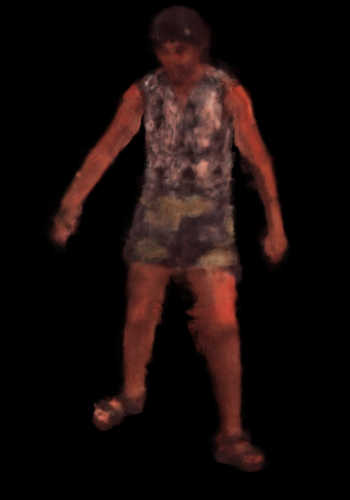}& 
        \includegraphics[width=\lw cm]{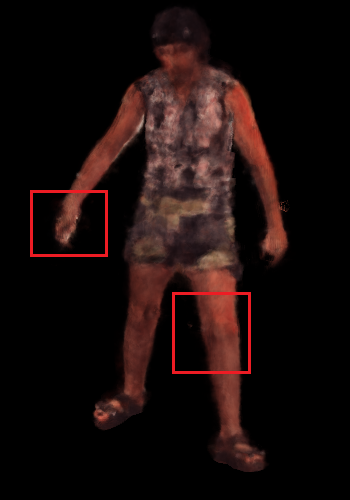}&
        \includegraphics[width=\lw cm]{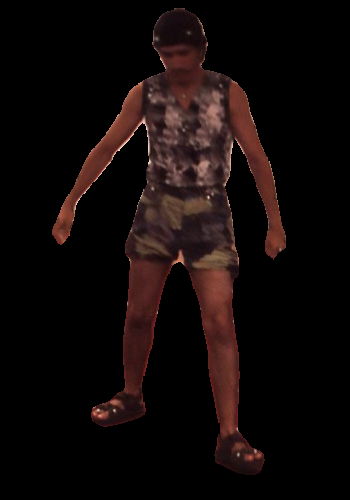}\\
        
        \small{(a) Neural Texture} & \small{(b) NHR} & \small{(c) D-NeRF} & \small{(d) AN} & \small{(e) ours} & \small{(f) Groundtruth}\\
\end{tabular}
}
\caption{Visualization of novel view reconstruction for the three test examples in H36M dataset. AN stands for Animatable NeRF. We highlight the significant improvements within boxes in our reconstructions.}
\label{fig:vis}
\end{figure*}

\textit{\textbf{Results on ZJU-MoCap.}} In addition to the results on the H36M dataset, we also show our average results for PSNR and SSIM for the ZJU-MoCap dataset in Table~\ref{tab:zju}. For the numerical results, we have similar PSNR and SSIM compared with NeuralBody and HumanNeRF on the novel view setting and best performance on novel pose synthesis. Compared with Animatable NeRF, we achieve a $19.1\%$  relative improvement compared to PSNR on novel views. Compared with the H36M dataset, ZJU-MoCap is larger, but its action variations are comparatively fewer, resulting in the improvements being smaller but still consistent.

{\textit{\textbf{Ablation for Covariance Loss.}} To assess the covariance loss $L_{cov}$ in Eq.~\ref{eq:covloss}, we replace it with three variations: 1) No extra loss for splitting, 2) Correlation loss $L_{corr}$, and 3) KL Divergence $L_{KLD}$ with the $N(0,1)$ distribution. $L_{corr}$ is the product of the correlation scores between every two dimensionalities in $\Psi$ following 
\begin{equation}
    L_{corr} = \frac{\sum (x_1 - \Bar{x_1})(x_2 - \Bar{x_2})...(x_n - \Bar{x_n})}{\sqrt{\sum (x_1 - \Bar{x_1})^2\sum (x_2 - \Bar{x_2})^2...\sum (x_n - \Bar{x_n})^2}}
\end{equation}
where $x_i$ is the $i^{th}$ dimension of training features.

We show the results in Table~\ref{tab:ablaloss} on the s1p subject of the H36M dataset. We note that, without any extra loss, our network already outperforms the original Animatable NeRF, indicating the constant features help find the appearance constancy in the sequence for rendering. In addition, all three methods have better results than the one with no additional loss applied, while Covariance loss in Eq.~\ref{eq:covloss} has the best performance. Using product instead of sum and the lack of examples compared with the feature dimensionality make $L_{corr}$ and $L_{KLD}$ less likely to be optimized. In contrast, $L_{cov}$ is capable of finding a stable solution to push $\psi^u$ to find the uniqueness for each frame. }

\textit{\textbf{Ablation for $G(\cdot)$.}} Instead of using Tx$^2$Former for fusing frame-unique feature, we show the ablations of different variations in Table~\ref{tab:ablaTx}. We compare Tx$^2$Former (denotes as Tx$^2$) with 1) unprocessed frame-level latent $\psi_i^u$, 2) average pooling of all $\Psi^u$, 3) the original combination of two transformer layers $Tx$ and 4) replacing the first layer with average pooling. We also compare these methods on s1p object of the H36m dataset and show that Tx$^2$Former has the best performance, indicating Tx$^2$Former being able to mine the neighbor features and find out the missing features in $\psi^u_i$.

\subsection{Visualization Results}\label{visr}
\textbf{\textit{Comparison with other methods.}} We show some rendered images in Figure~\ref{fig:vis} on the H36M dataset with novel view reconstruction results. In addition to the Neural Texture \cite{thies2019deferred}, NHR \cite{wu2020multi} and Animatable NeRF \cite{peng2021animatable}, we also compared with D-NeRF \cite{pumarola2021d} for reconstruction. D-NeRF utilizes the SE (3) transformations to find the corresponding points in the canonical space for the points observed.

\begin{table}[t]
\centering
\def\lw{1}
\def\lf{2}
\def\ls{0.05}
\resizebox{\linewidth}{!}
{
\begin{tabular}{p{1cm}<{\centering}p{1.2cm}<{\centering}p{\lw cm}<{\centering}p{\lw cm}<{\centering}p{\lw cm}<{\centering}p{\lw cm}<{\centering}p{\lw cm}<{\centering}p{\lw cm}<{\centering}p{\lw cm}<{\centering}} 
\toprule
Metric & Setting  & Rand & NT & NHR & NB & HN & AN & Ours\\
\midrule
\multirow{2}{*}{PSNR} 
& View & 17.83 & 22.61 & 23.25 & 28.90 & \textbf{29.01} & 27.10 & 28.87 \\
& Pose & 18.19 & 21.55 & 21.88 & 23.06 & 23.20 & 23.16 & \textbf{23.62} \\
\midrule
\multirow{2}{*}{SSIM} 
& View & 0.801 & 0.899 & 0.905 & 0.967 & \textbf{0.966} & 0.949 & 0.955 \\
& Pose & 0.790 & 0.860 & 0.863 & 0.879 & 0.885 & 0.893 & \textbf{0.899} \\
\bottomrule
\end{tabular}
}
\smallskip
\caption{Results of PSNR and SSIM for novel view and pose synthesis on ZJU-MoCap dataset. NT, HN and AN represents Neural Textures, HumanNeRF and Animatable NeRF. `Pose' and `View' represent the novel pose and novel view settings respectively.} 
\label{tab:zju}
\end{table}

\begin{table}[t]
\centering
\def\lw{1.3}
\def\ls{0.05}
\resizebox{\linewidth}{!}
{
\begin{tabular}{p{1.8cm}<{\centering}p{\ls cm}p{\lw cm}<{\centering}p{\lw cm}<{\centering}p{\lw cm}<{\centering}p{\lw cm}<{\centering}p{\lw cm}<{\centering}} 
\toprule
Loss Type && AN &  No Loss & $L_{corr}$ & $L_{KLD}$ & $L_{cov}$ \\
\midrule
PSNR && 22.05 &  22.86 & 23.52 & 23.67 & \textbf{24.52} \\
\bottomrule
\smallskip
\end{tabular}
}
\caption{{Ablation for using different loss functions to replace $L_{cov}$ in Eq.~\ref{eq:finalloss}, along with original Animatable NeRF (denotes as AN) results. All methods except AN are results for our model. `No Loss' is to remove $L_{cov}$ in $L$ of Eq.~\ref{eq:finalloss}.} }
\label{tab:ablaloss}
\end{table}

\begin{table}[t]
\centering
\def\lw{1.3}
\def\ls{0.05}
\resizebox{\linewidth}{!}
{
\begin{tabular}{p{1.8cm}<{\centering}p{\ls cm}p{\lw cm}<{\centering}p{\lw cm}<{\centering}p{\lw cm}<{\centering}p{\lw cm}<{\centering}p{\lw cm}<{\centering}} 
\toprule
Loss Type && $\psi_i^l$ & $Avg$ &  $Tx$ & $Avg + T_2$  & $Tx^2$ \\
\midrule
PSNR && 22.05 &  23.09 & 23.19 & 23.82 &\textbf{ 24.52} \\
\bottomrule
\smallskip
\end{tabular}
}
\caption{Ablation for different frame-unique features selection as $G(\cdot)$. $Avg$ is average pooling and $Tx$ is one-layer encoder.}
\label{tab:ablaTx}
\end{table}

\begin{figure*}[h]
\centering
\def\lw{2.5}
\def\ls{0.2}
\resizebox{0.9\linewidth}{!}
{
\centering
\begin{tabular}{p{\lw cm}<{\centering}p{\lw cm}<{\centering}p{\lw cm}<{\centering}p{\ls cm}p{\lw cm}<{\centering}p{\lw cm}<{\centering}p{\lw cm}<{\centering}}
        \\
        
        \includegraphics[width=\lw cm]{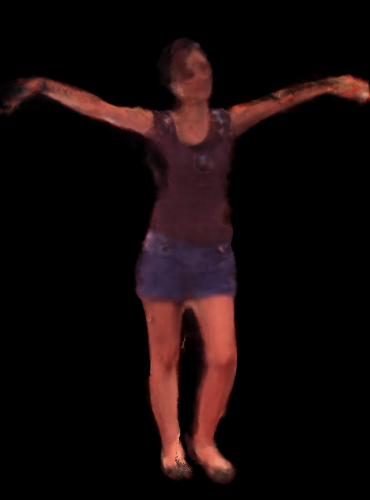}&
        \includegraphics[width=\lw cm]{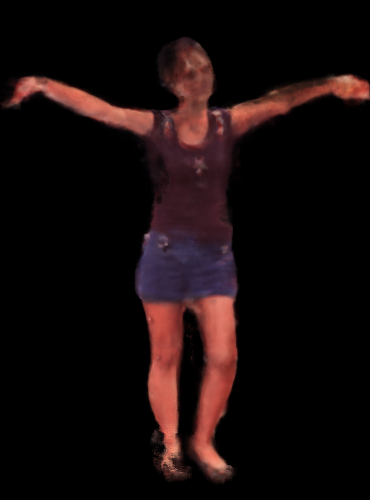}&
        \includegraphics[width=\lw cm]{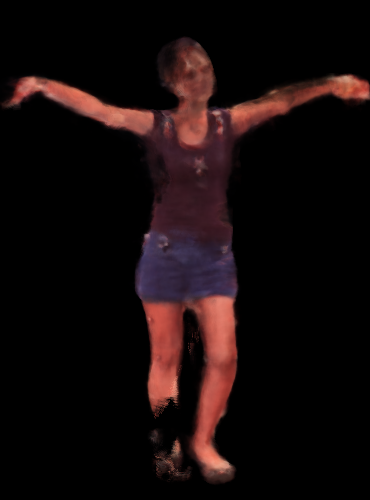}&&
        \includegraphics[width=\lw cm]{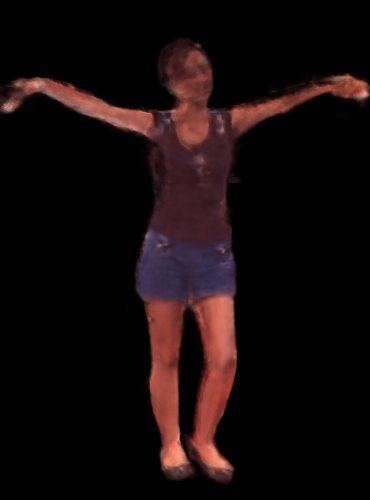}&
        \includegraphics[width=\lw cm]{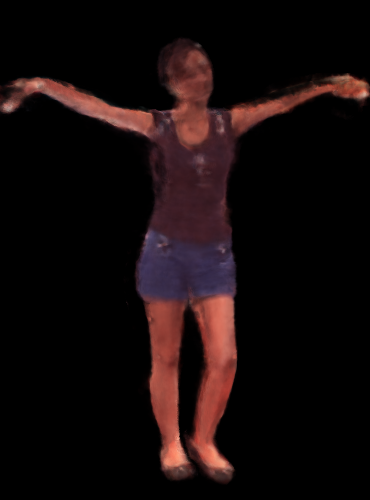}&
        \includegraphics[width=\lw cm]{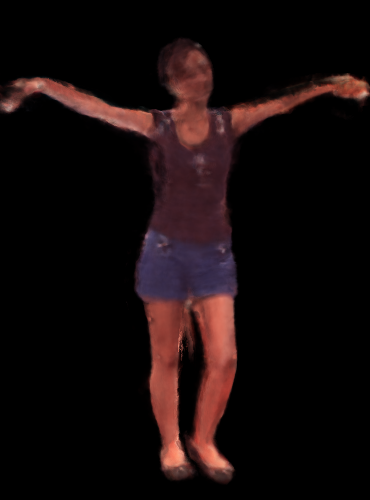}\\
        \includegraphics[width=\lw cm]{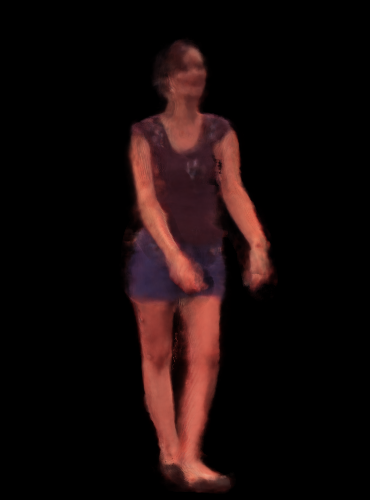}&
        \includegraphics[width=\lw cm]{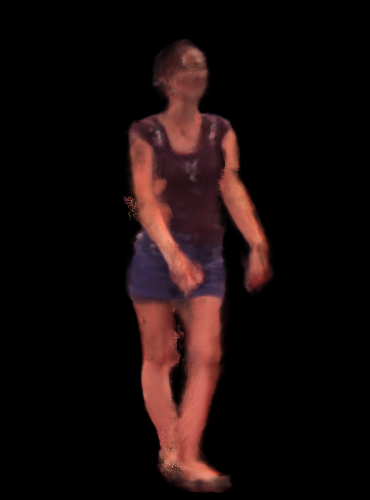}&
        \includegraphics[width=\lw cm]{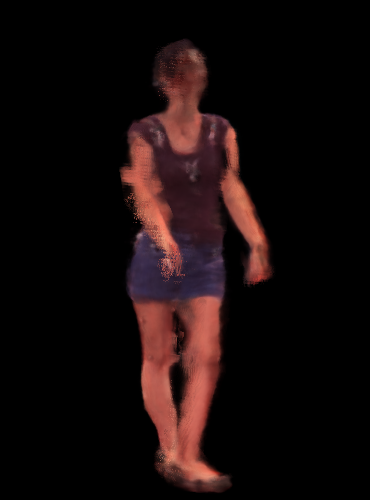}&&
        \includegraphics[width=\lw cm]{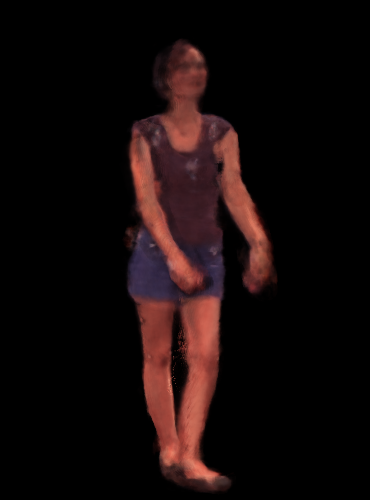}&
        \includegraphics[width=\lw cm]{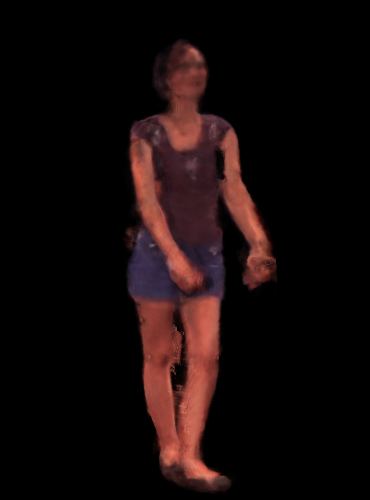}&
        \includegraphics[width=\lw cm]{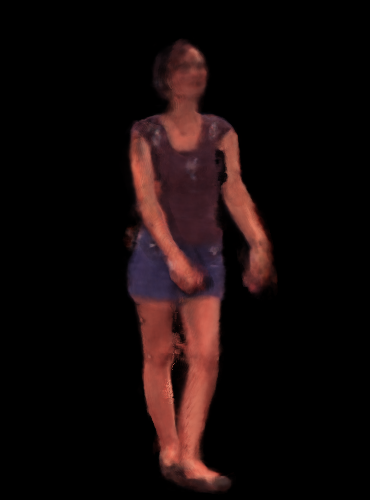}\\
        \small{(a) $\Delta t=0$} & \small{(b) $\Delta t=4$} & \small{(c) $\Delta t=8$} && \small{(a) $\Delta t=0$} & \small{(b) $\Delta t=4$} & \small{(c) $\Delta t=8$}\\
        \multicolumn{3}{c}{\small{Animatable NeRF}} && \multicolumn{3}{c}{\small{Ours}}\\
\end{tabular}
}
\caption{Visualization of novel view reconstruction for the test examples in H36M dataset with neural blend weight field from other frames. $\Delta t$ indicates the differences of the frame id in the temporal sequence.}
\label{fig:abla}
\end{figure*}

With the introduction of decoupling of the constant and frame-unique appearance with the blend weight between the statistic shape and the final reconstructed shape, our model can capture the body shape more accurately and with clearer boundaries. 
For example, in the first and third rows, our method is capable of making the correct prediction for the position of the two legs and having a clearer facial appearance. We have clearer rendered clothing results for the second example than Animated NeRF. Using the constant features to capture such appearance constancy allows our network to find and preserve the temporal similarities instead of being dominated by unique patterns.

\textit{\textbf{Evaluation for Appearance Constancy.}} Since we separate the temporal constant and framewise-unique features for dynamic human body shape rendering, in this experiment, we assess how much information has been captured by the constant vector as the appearance constancy. We exchange the frame-unique representations $\psi^u$ between different frames to see how big a difference the rendered image has over the original rendered results. We select two frames each time, one from $t$ and another from $t+\Delta t$, and apply the neural blend weight field from frame $t+\Delta t$ to frame $t$. The images of the person are captured with the same camera viewpoint but from different timestamps. If the constant feature captures enough appearance constancy, changes in the frame-unique feature vector will not have big differences in the final rendered image since the temporal constant feature is shared between two frames.

\CHD{
We show the results in Figure~\ref{fig:abla}. With the framewise features from a different timestamp, Animatable NeRF reconstructs a blurred image with general appearances unchanged, indicating most patterns and appearances are similar between the frames. As a comparison, CAT-NeRF stores such appearance constancies in the sequence. For example, in the second row, when we use the framewise features from other frames, the facial appearance of Animatable NeRF is not stable, while ours show a more accurate rendering. When such appearances, such as facial patterns, are constant and shared across all the frames, our model can construct a more confident result from the sequential inputs.
}

\section{Conclusion}
\CHD{
We introduce CAT-NeRF for mining and separating the appearance constancy and uniqueness for dynamic body shape rendering. The constant feature predicts the constant appearances that are shareable for all frames for the same person in the sequence, while the unique framewise feature simulates the unique dynamics in the sequential rendering. In addition, we introduce Tx$^2$Former for combining different levels of features and a specific Covariance Loss to ensure the unique appearance for each frame is correctly captured. Our method outperforms state-of-the-art methods on the public datasets, H36M \cite{ionescu2013human3} and ZJU-MoCap \cite{peng2021neural}.
}

\textbf{\textit{Acknowledgement.}} This research is based upon work supported in part by the Office of the Director of National Intelligence (ODNI), Intelligence Advanced Research Projects Activity (IARPA), via [2022-21102100007]. The views and conclusions contained herein are those of the authors and should not be interpreted as necessarily representing the official policies, either expressed or implied, of ODNI, IARPA, or the U.S. Government. The U.S. Government is authorized to reproduce and distribute reprints for governmental purposes notwithstanding any copyright annotation therein.

{\small
\bibliographystyle{ieee_fullname}
\bibliography{egmain}

\begin{thebibliography}{10}\itemsep=-1pt

\bibitem{aliev2020neural}
Kara-Ali Aliev, Artem Sevastopolsky, Maria Kolos, Dmitry Ulyanov, and Victor
  Lempitsky.
\newblock Neural point-based graphics.
\newblock In {\em ECCV}, pages 696--712. Springer, 2020.

\bibitem{bhatnagar2020loopreg}
Bharat~Lal Bhatnagar, Cristian Sminchisescu, Christian Theobalt, and Gerard
  Pons-Moll.
\newblock Loopreg: Self-supervised learning of implicit surface
  correspondences, pose and shape for 3d human mesh registration.
\newblock {\em NeurIPS}, 33:12909--12922, 2020.

\bibitem{bogo2016keep}
Federica Bogo, Angjoo Kanazawa, Christoph Lassner, Peter Gehler, Javier Romero,
  and Michael~J Black.
\newblock Keep it smpl: Automatic estimation of 3d human pose and shape from a
  single image.
\newblock In {\em ECCV}, pages 561--578, 2016.

\bibitem{chen2021snarf}
Xu Chen, Yufeng Zheng, Michael~J Black, Otmar Hilliges, and Andreas Geiger.
\newblock Snarf: Differentiable forward skinning for animating non-rigid neural
  implicit shapes.
\newblock In {\em ICCV}, pages 11594--11604, 2021.

\bibitem{collet2015high}
Alvaro Collet, Ming Chuang, Pat Sweeney, Don Gillett, Dennis Evseev, David
  Calabrese, Hugues Hoppe, Adam Kirk, and Steve Sullivan.
\newblock High-quality streamable free-viewpoint video.
\newblock {\em TOG}, 34(4):1--13, 2015.

\bibitem{debevec2000acquiring}
Paul Debevec, Tim Hawkins, Chris Tchou, Haarm-Pieter Duiker, Westley Sarokin,
  and Mark Sagar.
\newblock Acquiring the reflectance field of a human face.
\newblock In {\em SIGGRAPH}, pages 145--156, 2000.

\bibitem{dong2020motion}
Junting Dong, Qing Shuai, Yuanqing Zhang, Xian Liu, Xiaowei Zhou, and Hujun
  Bao.
\newblock Motion capture from internet videos.
\newblock In {\em ECCV}, pages 210--227. Springer, 2020.

\bibitem{dosovitskiy2020image}
Alexey Dosovitskiy, Lucas Beyer, Alexander Kolesnikov, Dirk Weissenborn,
  Xiaohua Zhai, Thomas Unterthiner, Mostafa Dehghani, Matthias Minderer, Georg
  Heigold, Sylvain Gelly, et~al.
\newblock An image is worth 16x16 words: Transformers for image recognition at
  scale.
\newblock {\em arXiv preprint arXiv:2010.11929}, 2020.

\bibitem{dou2016fusion4d}
Mingsong Dou, Sameh Khamis, Yury Degtyarev, Philip Davidson, Sean~Ryan Fanello,
  Adarsh Kowdle, Sergio~Orts Escolano, Christoph Rhemann, David Kim, Jonathan
  Taylor, et~al.
\newblock Fusion4d: Real-time performance capture of challenging scenes.
\newblock {\em ACM ToG}, 35(4):1--13, 2016.

\bibitem{guo2019relightables}
Kaiwen Guo, Peter Lincoln, Philip Davidson, Jay Busch, Xueming Yu, Matt Whalen,
  Geoff Harvey, Sergio Orts-Escolano, Rohit Pandey, Jason Dourgarian, et~al.
\newblock The relightables: Volumetric performance capture of humans with
  realistic relighting.
\newblock {\em ACM TOG}, 38:1--19, 2019.

\bibitem{hedman2021baking}
Peter Hedman, Pratul~P Srinivasan, Ben Mildenhall, Jonathan~T Barron, and Paul
  Debevec.
\newblock Baking neural radiance fields for real-time view synthesis.
\newblock In {\em ICCV}, pages 5875--5884, 2021.

\bibitem{huang2020arch}
Zeng Huang, Yuanlu Xu, Christoph Lassner, Hao Li, and Tony Tung.
\newblock Arch: Animatable reconstruction of clothed humans.
\newblock In {\em CVPR}, pages 3093--3102, 2020.

\bibitem{ionescu2013human3}
Catalin Ionescu, Dragos Papava, Vlad Olaru, and Cristian Sminchisescu.
\newblock Human3. 6m: Large scale datasets and predictive methods for 3d human
  sensing in natural environments.
\newblock {\em TPAMI}, pages 1325--1339, 2013.

\bibitem{jiang2020coherent}
Wen Jiang, Nikos Kolotouros, Georgios Pavlakos, Xiaowei Zhou, and Kostas
  Daniilidis.
\newblock Coherent reconstruction of multiple humans from a single image.
\newblock In {\em CVPR}, pages 5579--5588, 2020.

\bibitem{joo2018total}
Hanbyul Joo, Tomas Simon, and Yaser Sheikh.
\newblock Total capture: A 3d deformation model for tracking faces, hands, and
  bodies.
\newblock In {\em CVPR}, pages 8320--8329, 2018.

\bibitem{kanazawa2018end}
Angjoo Kanazawa, Michael~J Black, David~W Jacobs, and Jitendra Malik.
\newblock End-to-end recovery of human shape and pose.
\newblock In {\em CVPR}, pages 7122--7131, 2018.

\bibitem{kingma2014adam}
Diederik~P Kingma and Jimmy Ba.
\newblock Adam: A method for stochastic optimization.
\newblock {\em arXiv preprint arXiv:1412.6980}, 2014.

\bibitem{li2022tava}
Ruilong Li, Julian Tanke, Minh Vo, Michael Zollhofer, Jurgen Gall, Angjoo
  Kanazawa, and Christoph Lassner.
\newblock Tava: Template-free animatable volumetric actors.
\newblock {\em arXiv preprint arXiv:2206.08929}, 2022.

\bibitem{li2021neural}
Zhengqi Li, Simon Niklaus, Noah Snavely, and Oliver Wang.
\newblock Neural scene flow fields for space-time view synthesis of dynamic
  scenes.
\newblock In {\em CVPR}, pages 6498--6508, 2021.

\bibitem{liu2021neuralactor}
Lingjie Liu, Marc Habermann, Viktor Rudnev, Kripasindhu Sarkar, Jiatao Gu, and
  Christian Theobalt.
\newblock Neural actor: Neural free-view synthesis of human actors with pose
  control.
\newblock {\em ACM SIGGRAPH Asia}, 2021.

\bibitem{liu2021swin}
Ze Liu, Yutong Lin, Yue Cao, Han Hu, Yixuan Wei, Zheng Zhang, Stephen Lin, and
  Baining Guo.
\newblock Swin transformer: Hierarchical vision transformer using shifted
  windows.
\newblock In {\em ICCV}, pages 10012--10022, 2021.

\bibitem{loper2015smpl}
Matthew Loper, Naureen Mahmood, Javier Romero, Gerard Pons-Moll, and Michael~J
  Black.
\newblock Smpl: A skinned multi-person linear model.
\newblock {\em TOG}, 34(6):1--16, 2015.

\bibitem{martin2021nerf}
Ricardo Martin-Brualla, Noha Radwan, Mehdi~SM Sajjadi, Jonathan~T Barron,
  Alexey Dosovitskiy, and Daniel Duckworth.
\newblock Nerf in the wild: Neural radiance fields for unconstrained photo
  collections.
\newblock In {\em CVPR}, pages 7210--7219, 2021.

\bibitem{mildenhall2020nerf}
Ben Mildenhall, Pratul~P Srinivasan, Matthew Tancik, Jonathan~T Barron, Ravi
  Ramamoorthi, and Ren Ng.
\newblock Nerf: Representing scenes as neural radiance fields for view
  synthesis.
\newblock In {\em ECCV}, pages 405--421. Springer, 2020.

\bibitem{natsume2019siclope}
Ryota Natsume, Shunsuke Saito, Zeng Huang, Weikai Chen, Chongyang Ma, Hao Li,
  and Shigeo Morishima.
\newblock Siclope: Silhouette-based clothed people.
\newblock In {\em CVPR}, pages 4480--4490, 2019.

\bibitem{niemeyer2020differentiable}
Michael Niemeyer, Lars Mescheder, Michael Oechsle, and Andreas Geiger.
\newblock Differentiable volumetric rendering: Learning implicit 3d
  representations without 3d supervision.
\newblock In {\em CVPR}, pages 3504--3515, 2020.

\bibitem{noguchi2021neural}
Atsuhiro Noguchi, Xiao Sun, Stephen Lin, and Tatsuya Harada.
\newblock Neural articulated radiance field.
\newblock In {\em ICCV}, pages 5762--5772, 2021.

\bibitem{osman2020star}
Ahmed~AA Osman, Timo Bolkart, and Michael~J Black.
\newblock Star: Sparse trained articulated human body regressor.
\newblock In {\em ECCV}, pages 598--613, 2020.

\bibitem{park2021nerfies}
Keunhong Park, Utkarsh Sinha, Jonathan~T Barron, Sofien Bouaziz, Dan~B Goldman,
  Steven~M Seitz, and Ricardo Martin-Brualla.
\newblock Nerfies: Deformable neural radiance fields.
\newblock In {\em ICCV}, pages 5865--5874, 2021.

\bibitem{park2021hypernerf}
Keunhong Park, Utkarsh Sinha, Peter Hedman, Jonathan~T Barron, Sofien Bouaziz,
  Dan~B Goldman, Ricardo Martin-Brualla, and Steven~M Seitz.
\newblock Hypernerf: A higher-dimensional representation for topologically
  varying neural radiance fields.
\newblock {\em arXiv preprint arXiv:2106.13228}, 2021.

\bibitem{pavlakos2019expressive}
Georgios Pavlakos, Vasileios Choutas, Nima Ghorbani, Timo Bolkart, Ahmed~AA
  Osman, Dimitrios Tzionas, and Michael~J Black.
\newblock Expressive body capture: 3d hands, face, and body from a single
  image.
\newblock In {\em CVPR}, pages 10975--10985, 2019.

\bibitem{peng2021animatable}
Sida Peng, Junting Dong, Qianqian Wang, Shangzhan Zhang, Qing Shuai, Xiaowei
  Zhou, and Hujun Bao.
\newblock Animatable neural radiance fields for modeling dynamic human bodies.
\newblock In {\em ICCV}, pages 14314--14323, 2021.

\bibitem{peng2021neural}
Sida Peng, Yuanqing Zhang, Yinghao Xu, Qianqian Wang, Qing Shuai, Hujun Bao,
  and Xiaowei Zhou.
\newblock Neural body: Implicit neural representations with structured latent
  codes for novel view synthesis of dynamic humans.
\newblock In {\em CVPR}, pages 9054--9063, 2021.

\bibitem{prokudin2021smplpix}
Sergey Prokudin, Michael~J Black, and Javier Romero.
\newblock Smplpix: Neural avatars from 3d human models.
\newblock In {\em WACV}, pages 1810--1819, 2021.

\bibitem{pumarola2021d}
Albert Pumarola, Enric Corona, Gerard Pons-Moll, and Francesc Moreno-Noguer.
\newblock D-nerf: Neural radiance fields for dynamic scenes.
\newblock In {\em CVPR}, pages 10318--10327, 2021.

\bibitem{raj2021anr}
Amit Raj, Julian Tanke, James Hays, Minh Vo, Carsten Stoll, and Christoph
  Lassner.
\newblock Anr: Articulated neural rendering for virtual avatars.
\newblock In {\em CVPR}, pages 3722--3731, 2021.

\bibitem{romero2017embodied}
Javier Romero, Dimitrios Tzionas, and Michael~J Black.
\newblock Embodied hands: Modeling and capturing hands and bodies together.
\newblock {\em TOG}, 36(6):1--17, 2017.

\bibitem{saito2019pifu}
Shunsuke Saito, Zeng Huang, Ryota Natsume, Shigeo Morishima, Angjoo Kanazawa,
  and Hao Li.
\newblock Pifu: Pixel-aligned implicit function for high-resolution clothed
  human digitization.
\newblock In {\em ICCV}, pages 2304--2314, 2019.

\bibitem{saito2020pifuhd}
Shunsuke Saito, Tomas Simon, Jason Saragih, and Hanbyul Joo.
\newblock Pifuhd: Multi-level pixel-aligned implicit function for
  high-resolution 3d human digitization.
\newblock In {\em CVPR}, pages 84--93, 2020.

\bibitem{shysheya2019textured}
Aliaksandra Shysheya, Egor Zakharov, Kara-Ali Aliev, Renat Bashirov, Egor
  Burkov, Karim Iskakov, Aleksei Ivakhnenko, Yury Malkov, Igor Pasechnik,
  Dmitry Ulyanov, et~al.
\newblock Textured neural avatars.
\newblock In {\em CVPR}, pages 2387--2397, 2019.

\bibitem{sitzmann2019scene}
Vincent Sitzmann, Michael Zollh{\"o}fer, and Gordon Wetzstein.
\newblock Scene representation networks: Continuous 3d-structure-aware neural
  scene representations.
\newblock {\em NeurIPS}, 32, 2019.

\bibitem{su2021nerf}
Shih-Yang Su, Frank Yu, Michael Zollh{\"o}fer, and Helge Rhodin.
\newblock A-nerf: Articulated neural radiance fields for learning human shape,
  appearance, and pose.
\newblock {\em NeurIPS}, 34:12278--12291, 2021.

\bibitem{su2020robustfusion}
Zhuo Su, Lan Xu, Zerong Zheng, Tao Yu, Yebin Liu, and Lu Fang.
\newblock Robustfusion: Human volumetric capture with data-driven visual cues
  using a rgbd camera.
\newblock In {\em ECCV}, pages 246--264, 2020.

\bibitem{thies2019deferred}
Justus Thies, Michael Zollh{\"o}fer, and Matthias Nie{\ss}ner.
\newblock Deferred neural rendering: Image synthesis using neural textures.
\newblock {\em TOG}, 38(4):1--12, 2019.

\bibitem{touvron2021training}
Hugo Touvron, Matthieu Cord, Matthijs Douze, Francisco Massa, Alexandre
  Sablayrolles, and Herv{\'e} J{\'e}gou.
\newblock Training data-efficient image transformers \& distillation through
  attention.
\newblock In {\em ICML}, pages 10347--10357, 2021.

\bibitem{vaswani2017attention}
Ashish Vaswani, Noam Shazeer, Niki Parmar, Jakob Uszkoreit, Llion Jones,
  Aidan~N Gomez, {\L}ukasz Kaiser, and Illia Polosukhin.
\newblock Attention is all you need.
\newblock {\em NeurIPS}, 30, 2017.

\bibitem{wang2022arah}
Shaofei Wang, Katja Schwarz, Andreas Geiger, and Siyu Tang.
\newblock Arah: Animatable volume rendering of articulated human sdfs.
\newblock In {\em ECCV}, 2022.

\bibitem{wang2004image}
Zhou Wang, Alan~C Bovik, Hamid~R Sheikh, and Eero~P Simoncelli.
\newblock Image quality assessment: from error visibility to structural
  similarity.
\newblock {\em TIP}, 13(4):600--612, 2004.

\bibitem{weng2022humannerf}
Chung-Yi Weng, Brian Curless, Pratul~P Srinivasan, Jonathan~T Barron, and Ira
  Kemelmacher-Shlizerman.
\newblock Humannerf: Free-viewpoint rendering of moving people from monocular
  video.
\newblock In {\em CVPR}, pages 16210--16220, 2022.

\bibitem{wu2020multi}
Minye Wu, Yuehao Wang, Qiang Hu, and Jingyi Yu.
\newblock Multi-view neural human rendering.
\newblock In {\em CVPR}, pages 1682--1691, 2020.

\bibitem{xiu2021icon}
Yuliang Xiu, Jinlong Yang, Dimitrios Tzionas, and Michael~J Black.
\newblock Icon: Implicit clothed humans obtained from normals.
\newblock {\em arXiv preprint arXiv:2112.09127}, 2021.

\bibitem{yariv2020multiview}
Lior Yariv, Yoni Kasten, Dror Moran, Meirav Galun, Matan Atzmon, Basri Ronen,
  and Yaron Lipman.
\newblock Multiview neural surface reconstruction by disentangling geometry and
  appearance.
\newblock {\em NeurIPS}, 33:2492--2502, 2020.

\bibitem{yoon2021pose}
Jae~Shin Yoon, Lingjie Liu, Vladislav Golyanik, Kripasindhu Sarkar, Hyun~Soo
  Park, and Christian Theobalt.
\newblock Pose-guided human animation from a single image in the wild.
\newblock In {\em CVPR}, pages 15039--15048, 2021.

\bibitem{zheng2019deephuman}
Zerong Zheng, Tao Yu, Yixuan Wei, Qionghai Dai, and Yebin Liu.
\newblock Deephuman: 3d human reconstruction from a single image.
\newblock In {\em ICCV}, pages 7739--7749, 2019.

\bibitem{zhuopen}
Haidong Zhu, Ye Yuan, Yiheng Zhu, Xiao Yang, and Ram Nevatia.
\newblock Open: Order-preserving pointcloud encoder decoder network for body
  shape refinement.
\newblock In {\em ICPR}, pages 521--527, 2022.

\bibitem{zhu2023gait}
Haidong Zhu, Zhaoheng Zheng, and Ram Nevatia.
\newblock Gait recognition using 3-d human body shape inference.
\newblock In {\em WACV}, pages 909--918, 2023.

\end{thebibliography}
}

\appendix
\null
\begin{table*}
\vskip .375in
\begin{center}
  {\Large \bf \papertitle \\\textit{Supplementary Material} \par}
\end{center}
\end{table*}

\newpage

In this supplementary material, we present some further background knowledge, experimental details, ablation studies, and visualization results that do not fit into the paper due to space limitations. 
We first introduce the background of NeRF, followed by the detailed network architecture description and the ablation study for the size of the latent feature and the number of views we used in the experiment. Finally, we present more visualization results on H36M and ZJU-MoCap datasets with novel views and poses.

\textit{\textbf{Review of Neural Radiance Field.}}
NeRF is a continuous volumetric field representing the density and color of the object or scene. For the color and density of a point for the viewing direction $d$, NeRF finds the ray $r$ that goes through the field and predicts the density $\sigma(x')$ and color value $c(x')$ for the 3-D point $x'$ on this ray following $\sigma(x'), c(x') = \mathbf{F}(x',d,l)$ using an MLP network $\mathbf{F}(\cdot)$. 
$l$ is a latent representation for the appearance of the object following \cite{martin2021nerf}. The final predicted color $\Tilde{C}(r)$ projected on the 2-D plain for the ray $r$ is accumulated with a random set of quadrature points $\{h_p\}^{m}_{p=1}\in [h_n, h_f]$ following \cite{mildenhall2020nerf}

\begin{equation}
  \Tilde{C}(r) = \sum_{p=1}^m exp(-\sum_{q=1}^{p-1}\ \sigma_q\delta_q) (1-exp(-\delta_p\sigma_p))c(p)  
\end{equation}
where $h_n$ and $h_f$ represent the near and far bound of the field for the sampled points. $\delta_p$ is the distance between two quadrature points $h_p$ and $h_{p+1}$. Due to the performance of under-fitting on high-frequency patterns, NeRF includes a position encoding $\gamma(x)$ for projecting the point $x$, normalized to $[-1,1]$, to the high dimension space following
\begin{equation}
\begin{aligned}
   \gamma(x) =(sin(2^0\pi x), cos(&2^0\pi x),...,\\ sin(&2^{L-1}\pi x), cos(2^{L-1}\pi x))
   \label{eq:cos}
\end{aligned}
\end{equation}
We follow \cite{mildenhall2020nerf} to use the three normalized coordinates of the point $x'$ with $L$ as 10 and three of the Cartesian viewing direction unit vector for $d$ with $L$ as 4.

\textit{\textbf{Network Architectures.}} In our experiment, we have five trainable modules in the proposed CAT-NeRF: neural radiance field, neural blend weight field, frame-unique field decoder, constant field decoder, and Tx$^2$Former $G(\cdot)$. 

For the neural radiance field, we follow \cite{peng2021animatable} to build an 11-layer MLP for decoding color and density. The dimensionality is set to 256 for the first 10 layers and 128 for the final layer. The positional encoding $\gamma(x)$ is sent to the first MLP layer and concatenated with the output of the $4^{th}$ layer as the input for $5^{th}$ MLP layer. The position encoding for the viewing direction  $\gamma(d)$ and appearance code $l$ is sent to the $9^{th}$ and $10^{th}$ concatenated with the previous layer's output. Density $\sigma$ and color $c$ are predicted from the $8^{th}$ and the last MLP layers, respectively. 

For the frame-unique and constant field decoders, we use two identical 2-layer MLP networks but do not share weights. The dimensionality for all layers is set to 64. For the neural blend weight field, we use an 8-layer MLP following \cite{peng2021animatable} where every layer is 128-d. We use the concatenation of the outputs from the constant and frame-unique field encoders along with $\gamma(x)$ as input for the first and fifth MLP layers. The final projection of $\Delta w$ is produced with the exponential output from the last layer. We use ReLU as the activation function following every MLP layer for all our networks except the final output layer. We set the dimensionality of features in Tx$^2$Former $G(\cdot)$ as 128 for both two layers and include one layer of Transformer encoder \cite{vaswani2017attention} for $T_1$ and $T_2$ in our experiment.

\textit{\textbf{Size of Latent Features.}} In our model, we introduce using the 128-dimension features for both frame-unique feature and constant feature representation. We show five different variations in the dimensionality for features representation in Table~\ref{tab:ablfeatdim}. Note that when we only have one 128-D feature vector for the frame-unique feature, the method defaults to Animatable NeRF. Compared with using the unique feature representation for every individual frame, we see that the constant feature alone boosts the performance for dynamic human body rendering and outperforms the original Animatable NeRF. When using the frame-unique feature along with the constant representation, the network shows the best PSNR value, while results for using 64-dim or 128-dim features do not change greatly.

\textit{\textbf{Number of Training Viewpoints.}} In addition to the main experiments where we use three training camera viewpoints for H36M, we also assess the performance and robustness of using fewer camera viewpoints comparing CAT-NeRF with Animatable NeRF. We train our network on one (viewpoint 0), two (viewpoint 0 and 1), and three (viewpoint 0,1 and 2) camera viewpoints, respectively, and report the PSNR results on the reconstruction of viewpoint 3 for inference.

\begin{figure*}[t]
\centering
\def\lw{2.5}
\def\ls{0.2}
\resizebox{0.95\linewidth}{!}
{
\centering
\begin{tabular}{p{\lw cm}<{\centering}p{\lw cm}<{\centering}p{\lw cm}<{\centering}p{\ls cm}p{\lw cm}<{\centering}p{\lw cm}<{\centering}p{\lw cm}<{\centering}}
        \\
        
        \includegraphics[width=\lw cm]{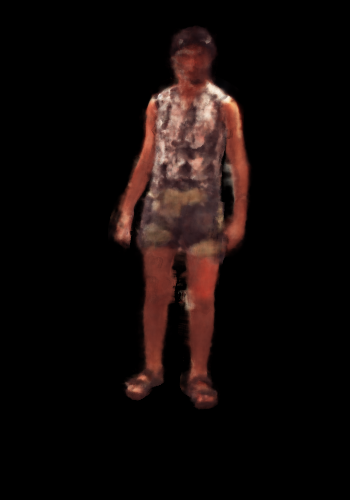}&
        \includegraphics[width=\lw cm]{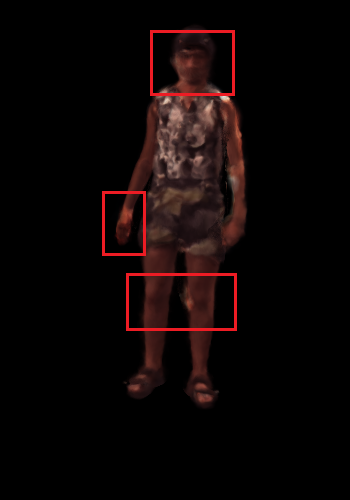}&
        \includegraphics[width=\lw cm]{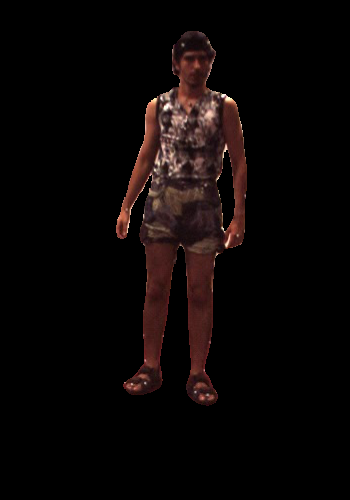}&&
        \includegraphics[width=\lw cm]{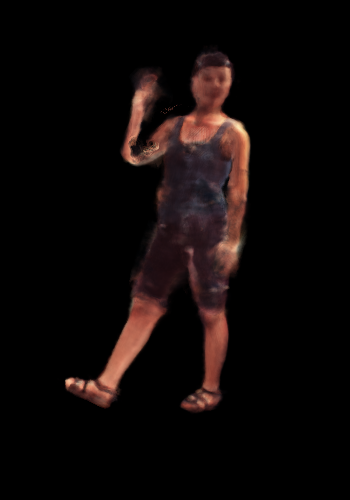}&
        \includegraphics[width=\lw cm]{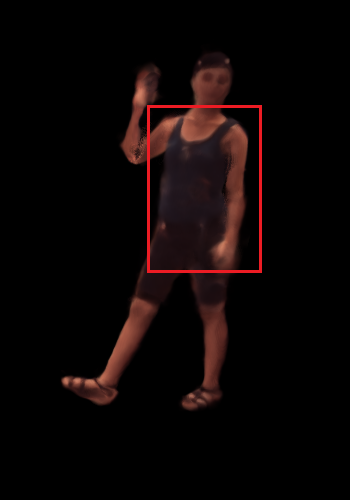}&
        \includegraphics[width=\lw cm]{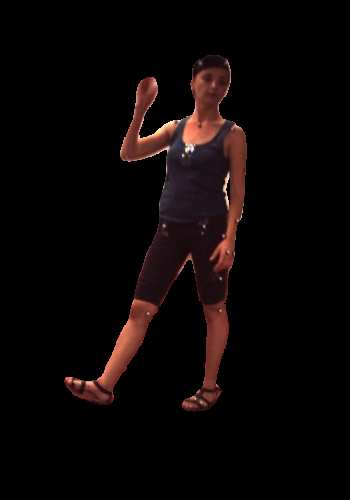}\\
        \includegraphics[width=\lw cm]{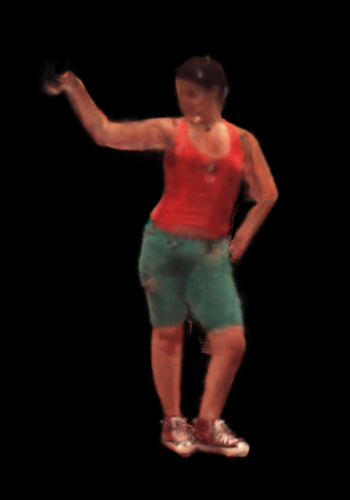}&
        \includegraphics[width=\lw cm]{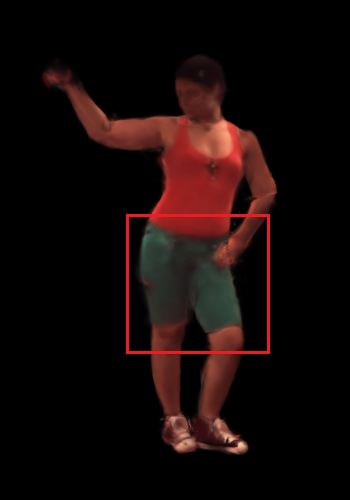}&
        \includegraphics[width=\lw cm]{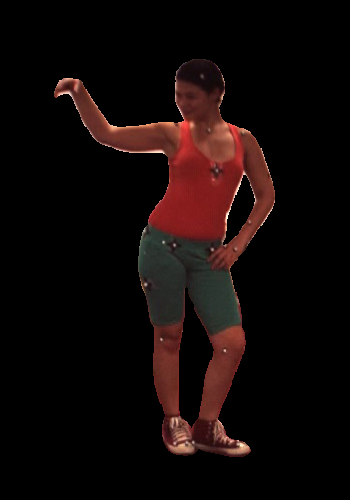}&&
        \includegraphics[width=\lw cm]{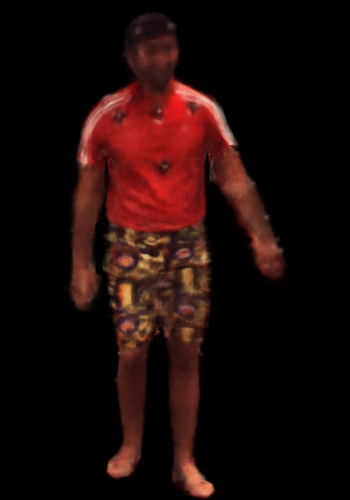}&
        \includegraphics[width=\lw cm]{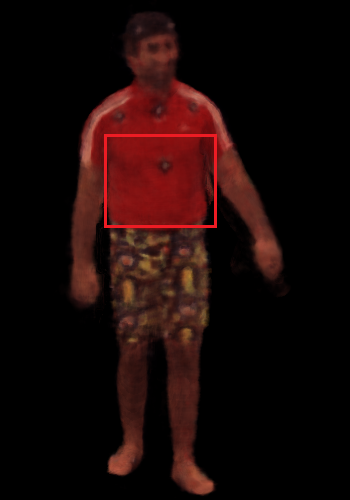}&
        \includegraphics[width=\lw cm]{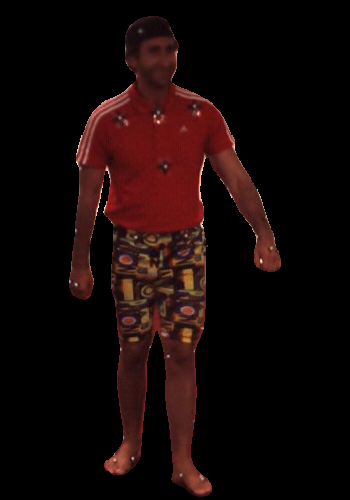}\\
        \includegraphics[width=\lw cm]{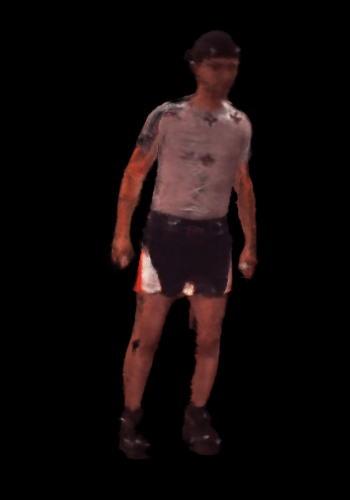}&
        \includegraphics[width=\lw cm]{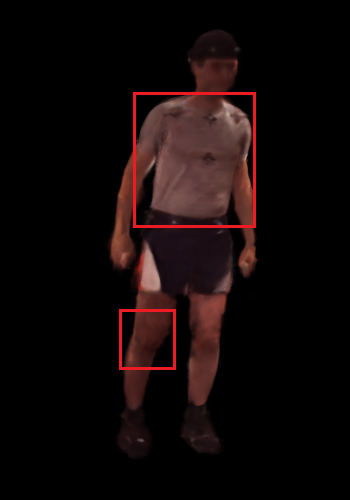}&
        \includegraphics[width=\lw cm]{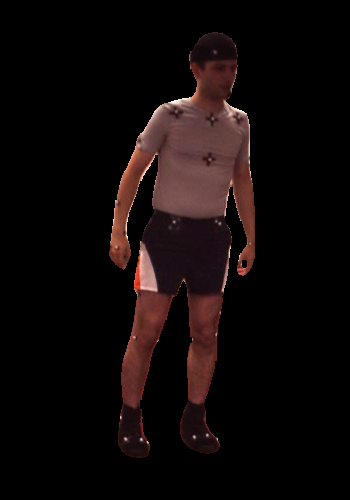}&&
        \includegraphics[width=\lw cm]{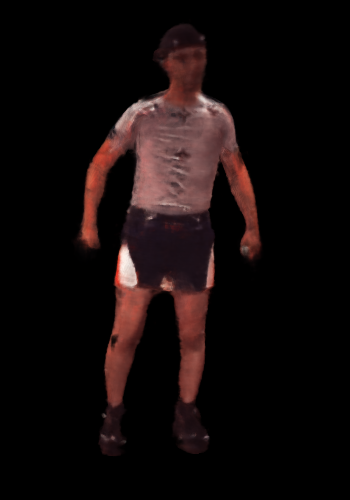}&
        \includegraphics[width=\lw cm]{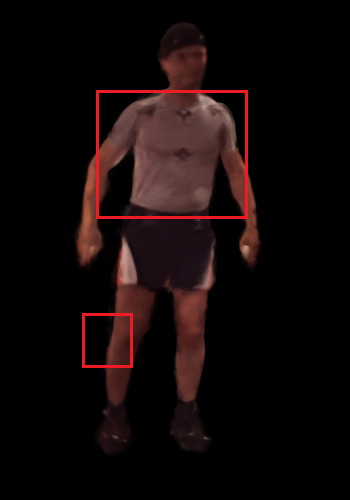}&
        \includegraphics[width=\lw cm]{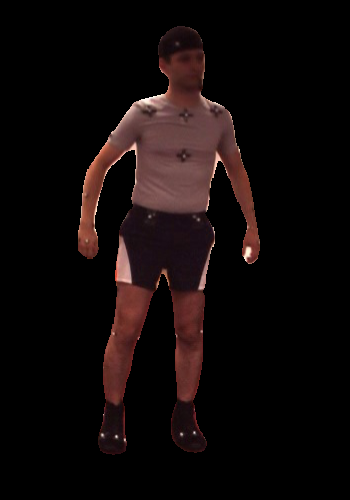}\\
        \small{(a) AN} & \small{(b) Ours} & \small{(c) Groundtruth} && \small{(a) AN} & \small{(b) Ours} & \small{(c) Groundtruth}\\
        \multicolumn{3}{c}{\small{Novel View}} && \multicolumn{3}{c}{\small{Novel Pose}}\\
\end{tabular}
}
\caption{Visualizations for novel pose and novel view on H36m datasets comparing (a) Animatable NeRF and (b) ours. Images in the column of AN are the outputs from Animatable NeRF.}
\label{fig:vish36m}
\end{figure*}

We show the results in Table~\ref{tab:ablvp}. With the maximum available viewpoints, both methods show the best performance for dynamic body shape rendering. When we reduce the available number of available viewpoints in the training set from 3 to 2, our proposed CAT-NeRF shows better consistency than Animatable NeRF, with only a 0.03 drop for PSNR, while Animatable NeRF drops 0.3. When the number of cameras is reduced to one, both methods fall significantly, while CAT-NeRF still achieves a PSNR value of 21.97, similar to the Animatable NeRF three-viewpoint result of 22.05. Mining the temporal constancy assists the model in extracting and extending the frame-level knowledge to video-level knowledge, helping the model achieve better performance even with only one training viewpoint.

\begin{table}[t]
\centering
\def\lw{.7}
\def\ls{0.01}
\resizebox{0.9\linewidth}{!}
{
\begin{tabular}{p{1.4cm}<{\centering}p{.2cm}<{\centering}p{\ls cm}p{\lw cm}<{\centering}p{\lw cm}<{\centering}p{\lw cm}<{\centering}p{\lw cm}<{\centering}p{\lw cm}<{\centering}} 
\toprule
\multirow{2}{*}{Feat. Dim} & $\psi^c$ && 0 & 64 & 128 & 64 & 128 \\
&$\psi^u$ && 128 & 64 & 128 & 128 & 0 \\
\midrule
\multicolumn{2}{c}{PSNR} && 22.05 & 24.16 & 24.52 & 24.35 & 23.14 \\
\bottomrule
\smallskip
\end{tabular}
}
\caption{Ablation results for different feature dimensions used by constant feature $\psi^c$ and frame-unique features $\psi^u$ of the network.} 
\label{tab:ablfeatdim}
\end{table}

\begin{table}[t]
\centering
\def\lw{1}
\def\ls{0.05}
\resizebox{.9\linewidth}{!}
{
\begin{tabular}{p{2.8cm}<{\centering}p{\ls cm}p{\lw cm}<{\centering}p{\lw cm}<{\centering}p{\lw cm}<{\centering}} 
\toprule
\# of Viewpoints && 1 & 2 & 3\\
\midrule
Animatable NeRF && 20.78 & 21.75 & 22.05 \\
Ours && 21.97 & 24.49 & 24.52 \\
\bottomrule
\smallskip
\end{tabular}
}
\caption{PSNR results for different numbers of training viewpoints compared with Animatable NeRF.} 
\label{tab:ablvp}
\end{table}

\begin{figure*}[t]
\centering
\def\lw{2.5}
\def\ls{0.2}
\resizebox{\linewidth}{!}
{

\centering
\def\lw{2.5}
\def\ls{0.2}
\resizebox{0.95\linewidth}{!}
{
\centering
\begin{tabular}{p{\lw cm}<{\centering}p{\lw cm}<{\centering}p{\lw cm}<{\centering}p{\ls cm}p{\lw cm}<{\centering}p{\lw cm}<{\centering}p{\lw cm}<{\centering}}
        \\

        \includegraphics[width=\lw cm]{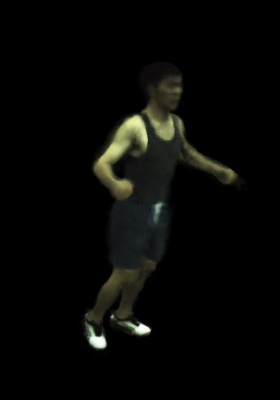}&
        \includegraphics[width=\lw cm]{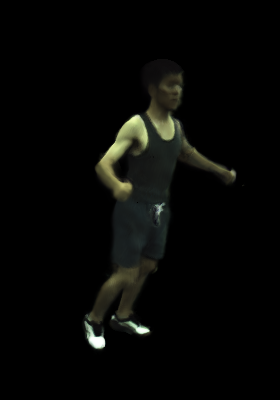}&
        \includegraphics[width=\lw cm]{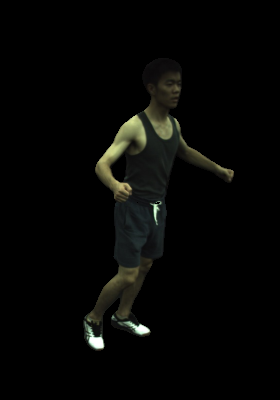}&&
        \includegraphics[width=\lw cm]{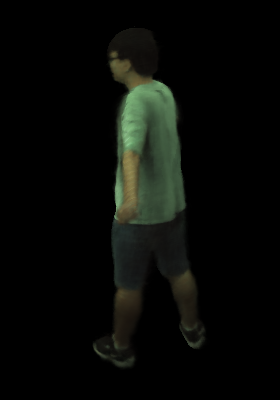}&
        \includegraphics[width=\lw cm]{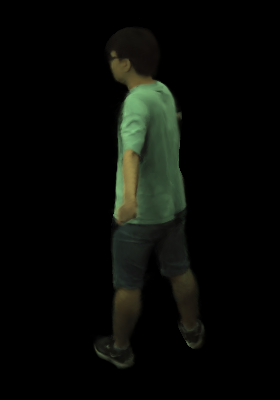}&
        \includegraphics[width=\lw cm]{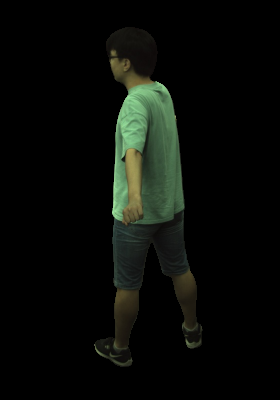}\\
        \includegraphics[width=\lw cm]{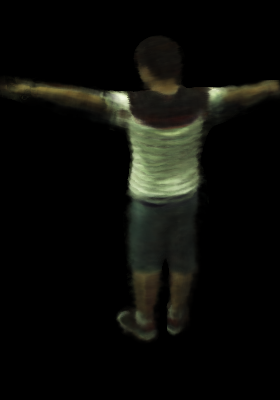}&
        \includegraphics[width=\lw cm]{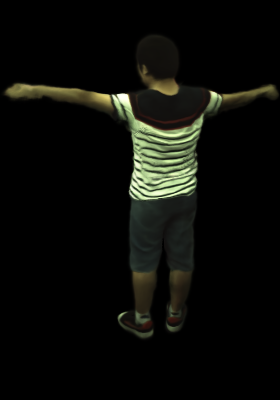}&
        \includegraphics[width=\lw cm]{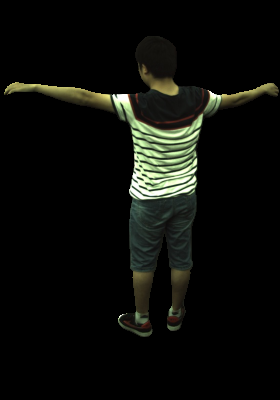}&&
        \includegraphics[width=\lw cm]{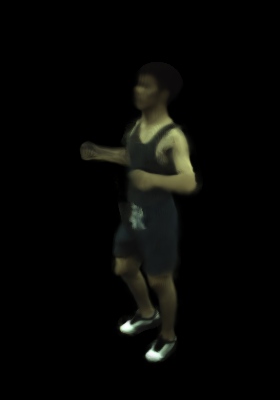}&
        \includegraphics[width=\lw cm]{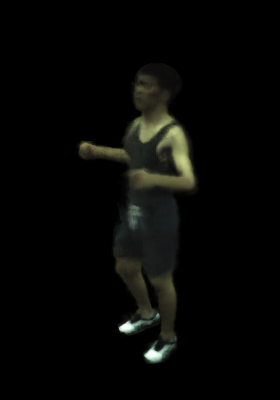}&
        \includegraphics[width=\lw cm]{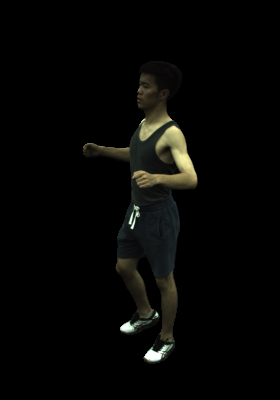}\\
        \small{(a) AN} & \small{(b) Ours} & \small{(c) Groundtruth} && \small{(a) AN} & \small{(b) Ours} & \small{(c) Groundtruth}\\
\end{tabular}
}
}
\caption{Visualizations for novel pose and novel view on ZJU-MoCap dataset comparing (a) Animatable NeRF and (b) ours. Images in the column of AN are the outputs from Animatable NeRF. }
\label{fig:viszjub}
\end{figure*}

\textit{\textbf{Visualization Results on H36M and ZJU-MoCap.}} In addition to the visualization results we show in our submission, we present more visualizations for both datasets on both novel view and novel pose settings compared with Animatable NeRF \cite{peng2021animatable}. We show the results for H36M in Figure~\ref{fig:vish36m} and ZJU-MoCap in Figure~\ref{fig:viszjub}. %

With CAT-NeRF, we see better-detailed reconstructions than Animatable NeRF on both datasets. For example, for both novel view and novel pose settings in the last row of  Figure~\ref{fig:vish36m}, Animatable NeRF introduces wrinkles that should not appear at the front of the person, while our method creates a more precise image. We can also observe similar differences in the bottom left example in Figure~\ref{fig:viszjub}, where Animatable NeRF fails to reconstruct the stripes on the shirt. In contrast, CAT-NeRF generates an image with a more precise boundary for the patterns and wrinkles on the clothes. In addition to the framewise results in the figures, we also attach a video for each dataset in the supplementary folder.

For the quality of predicted images of two datasets, we observe that, compared with H36M, visualization results on ZJU-MoCap are generally better. Since the number of samples used for training in H36M is smaller and pose differences between the frames are significant, H36M is a much harder dataset to render from a novel viewpoint or for novel poses compared with ZJU-MoCap. In addition, ZJU-MoCap has 4 camera viewpoints available during training, while H36M only has 3. In our submission, we present the visualization results from H36M to compare on a more challenging dataset for dynamic human body rendering.

\begin{figure}[t]
\centering
\def\lw{2.5}
\def\lwt{2.5}
\def\ls{0.2}
\resizebox{1\linewidth}{!}
{
\centering
\begin{tabular}{p{\lw cm}<{\centering}p{\lw cm}<{\centering}p{\lw cm}<{\centering}p{\lw cm}<{\centering}p{\lw cm}<{\centering}}
        \\
        \includegraphics[width=\lwt cm]{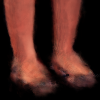}&
        \includegraphics[width=\lwt cm]{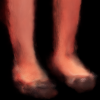}&
        \includegraphics[width=\lwt cm]{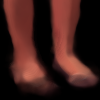}&
        \includegraphics[width=\lwt cm]{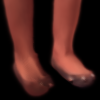}&
        \includegraphics[width=\lwt cm]{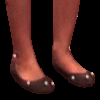}\\
        \includegraphics[width=\lwt cm]{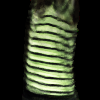}&
        \includegraphics[width=\lwt cm]{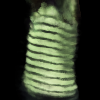}&
        \includegraphics[width=\lwt cm]{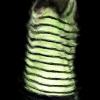}&
        \includegraphics[width=\lwt cm]{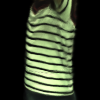}&
        \includegraphics[width=\lwt cm]{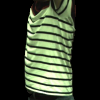}\\
        \small{(a) HumanNeRF} & \small{(b) TAVA} & \small{(c) ARAH} & \small{(d) Ours}& \small{(e) Groundtruth} \\
\end{tabular}
}
\caption{Visualization resutls for novel poses and novel views settings for examples in H36m and ZJU-MoCap dataset.}
\label{fig:reb}
\end{figure}

\textbf{Comparison with the Latest State-of-the-Art Methods:} We compared our method with some of the latest state-of-the-art methods such as HumanNeRF \cite{weng2022humannerf}, TAVA \cite{li2022tava}, and ARAH \cite{wang2022arah} as shown in Figure~\ref{fig:reb}. However, as most of these methods do not have results on H36m, the more complex dataset with fewer cameras and more action variations, and the experimental settings were not exactly the same, we rerun the experiments on the two public datasets used in our paper. We present qualitative comparisons on examples from H36m and ZJU-Mocap, where we zoom in on the corresponding body parts. Our approach achieves more precise boundaries of patterns such as stripes and shoes, with fewer artifacts such as wrinkles. Additionally, the key modules in our method are portable and can be applied to these latest state-of-the-art methods for further improvements.

\end{document}